\documentclass{article}
\PassOptionsToPackage{
  colorlinks=true,
  linkcolor=blue!70!black,
  citecolor=green!50!black,
  urlcolor=blue!70!black,
}{hyperref}

\usepackage[preprint]{temp}

\usepackage[utf8]{inputenc}
\usepackage[T1]{fontenc}
\usepackage{amsmath,amssymb,amsfonts,amsthm}
\usepackage{mathtools}
\usepackage{bm}

\newtheorem{proposition}{Proposition}
\newtheorem{definition}{Definition}

\usepackage{booktabs}
\usepackage{multirow}
\usepackage{graphicx}
\usepackage{subcaption}
\usepackage{float}

\usepackage{url}
\usepackage{cleveref}  
\usepackage{enumitem}
\usepackage{nicefrac}
\usepackage{xcolor}

\newcommand{\auroc}[1]{\text{AUROC}_{#1}}
\newcommand{\sd}[1]{${\scriptstyle\,\pm\,#1}$}

\title{Learning Transferable Predictability Representations}

\author{
  Diyali Goswami$^{1}$ \And Auroop R. Ganguly$^{1,2,3}$ \\[0.3cm]
  $^{1}$ Sustainability and Data Sciences Laboratory (SDS Lab), Northeastern University, Boston, MA, USA \\
  $^{2}$ AI4CaS: AI for Climate and Sustainability, Institute for Experiential AI, Northeastern University, Boston, MA, USA \\
  $^{3}$ Pacific Northwest National Laboratory (PNNL), Richland, WA, USA \\[0.2cm]
  \texttt{\{goswami.di, a.ganguly\}@northeastern.edu}
}

\begin{document}
\maketitle
\begingroup
\renewcommand\thefootnote{}
\footnotetext{Code will be released upon acceptance and repository cleanup.}
\addtocounter{footnote}{-1}
\endgroup

\begin{abstract}
We study the problem of assigning a scalar score to a short trajectory window that reflects its position on an ordered continuum of predictability regimes, spanning structured deterministic dynamics to unstructured stochastic noise. Existing methods address deterministic-versus-stochastic discrimination within a single system and do not produce scores with a consistent numerical interpretation across systems. We formalize this as ordinal estimation over a five-level predictability ladder and identify a structural source of cross-system ambiguity: ranking supervision alone leaves the score coordinate unfixed up to a monotone reparameterization, which we term the gauge freedom of ordinal scoring. We propose the Gauge-Fixed Ordinal Network (GON), a temporal convolutional model trained with an anchor-and-variance objective that pins level-wise score means to shared target coordinates. GON operates on 2-jet features that expose local trajectory geometry, preserved by smooth flows and disrupted by stochastic surrogate procedures. On five held-out dynamical systems, initializing from a pretrained GON checkpoint consistently outperforms training from scratch across all window budgets, with adaptation depth reflecting geometric proximity to the training family. Zero-shot scores retain ordinal structure at the stochastic boundary, where surrogate procedures most strongly disrupt nonlinear geometry, and pretrained initialization consistently beats scratch across all window budgets. Pairwise discrimination and globally coherent ordinal scoring are distinct properties requiring a stable score coordinate for cross-system transfer, with direct implications for predictability assessment, model selection, and early-warning diagnostics across natural and engineered dynamical systems.
\end{abstract}

\section{Introduction}
\label{sec:introduction}

An observed time series can often be viewed as a projection of an underlying dynamical system, though the generative process may be deterministic, stochastic, or a mixture of both. The source of predictive structure shapes which modeling strategies are appropriate. Deterministic chaotic systems are governed by an underlying flow that supports geometry-aware representations, even when long-horizon forecasting is limited by sensitive dependence on initial conditions. Purely stochastic processes are better described through statistical regularities. Both regimes can share low-order moments such as autocorrelation and power spectrum, making the distinction difficult to draw from observations alone. A signal may reflect low-dimensional deterministic dynamics, a stochastic process, or an intermediate regime that retains some structure while losing others. 

Classical tools approach this distinction within a single system. Surrogate hypothesis tests~\citep{theiler1992testing,schreiber1996improved} ask whether a signal departs from a chosen stochastic null. Lyapunov exponent estimators~\citep{wolf1985determining,rosenstein1993practical} quantify instability under an assumed deterministic model. Complexity statistics such as permutation entropy~\citep{bandt2002permutation} and sample entropy~\citep{richman2000physiological} compress a signal into a scalar intended to separate deterministic from stochastic behavior. Machine learning approaches have extended work on predictive structure in dynamical systems, including neural differential equation models~\citep{chen2018neural}, neural operators~\citep{li2020fourier,lu2021learning}, and pretrained models for chaotic forecasting~\citep{lai2025panda}. These methods characterize structure within a system or family, and a new signal typically requires new calibration, new surrogates, or a fresh thresholding decision. A detailed comparison is provided in Appendix~\ref{app:related}.

This paper addresses a specific problem: given a trajectory window, assign it a position on an ordered continuum of predictability regimes. We use predictability to mean the degree to which the underlying dynamics constrain future trajectory evolution, a property of the governing flow rather than of any particular model or observation process. We instantiate this continuum as a five-level \emph{predictability ladder} spanning structured deterministic dynamics, weakly chaotic regimes, strongly chaotic regimes, structured stochastic surrogates, and unstructured noise. Standard deterministic-versus-stochastic discrimination corresponds to one boundary in this richer ordinal problem. The central question is whether a learned ordinal score transfers across systems, rather than merely separating classes within a single one.

The ordinal setting introduces a structural ambiguity. If a scalar score correctly orders the ladder levels, any strictly increasing reparameterization preserves the same ordinal decisions after a corresponding threshold adjustment. Correct ordering alone does not determine the numerical coordinate of the score, an identifiability property of ordered regression models~\citep{mccullagh1980regression}, leaving scores across systems non-comparable even when both induce identical ordinal predictions. We refer to this as the \emph{gauge freedom} of ordinal scoring and identify it as the structural barrier to cross-system transfer; post-hoc per-system recalibration cannot substitute, because the intended use case is zero-shot transfer where no target-system data is available to fit a calibration mapping.

To address this, we introduce the \emph{Gauge-Fixed Ordinal Network} (GON), a temporal convolutional model trained with an anchor-based ordinal objective. The anchor fixes target locations for level-wise score means and penalizes excessive within-level spread, producing a scalar score with a stable coordinate convention. GON operates on 2-jet features (position, velocity, and acceleration), which expose local trajectory geometry~\citep{koenderink1987representation} preserved by smooth flows and disrupted by surrogate procedures; this representation has been applied to learned physical dynamics in~\citet{cranmer2020lagrangian}. We train on twelve dissipative chaotic ODE systems, chosen for their bounded attractors and varied geometric structure, and evaluate zero-shot transfer and few-shot adaptation on five fully held-out systems.

The main contributions are as follows.
\begin{enumerate}[leftmargin=*, itemsep=2pt]
    \item \textbf{Problem formulation.} We formalize ordinal predictability estimation and introduce the predictability ladder, identifying gauge freedom as the structural source of cross-system score ambiguity.
    \item \textbf{Method.} We propose GON, which combines 2-jet trajectory features with an anchor-based objective to learn a shared score coordinate across systems.
    \item \textbf{Empirical results.} We show that gauge fixing enables cross-system ordinal transfer: pretrained GON outperforms training from scratch across all held-out systems, zero-shot scores retain ordinal coherence at the surrogate boundary, and adaptation depth reflects geometric proximity to the training family.
\end{enumerate}

\section{Problem Formulation}
\label{sec:problem}
We formalize \emph{ordinal predictability estimation}: given a trajectory window, 
assign it a position on an ordered predictability spectrum that remains meaningful 
across systems, subsuming binary deterministic-versus-stochastic discrimination 
as a special case.

\subsection{The Predictability Ladder}
\label{subsec:ladder}

Let $x = (x_t)_{t=1}^T \in \mathbb{R}^{d \times T}$ denote a multivariate time series generated by an unknown process. We define a five-level ordered taxonomy
\[
\mathcal{R} = \{L_0, L_1, L_2, L_3, L_4\}
\]
called the \emph{predictability ladder}, arranged from more predictable to less predictable regimes.

\begin{definition}[Predictability Ladder]
\label{def:ladder}
The levels satisfy
\[
L_0 \prec L_1 \prec L_2 \prec L_3 \prec L_4,
\]
with the following interpretation:
\begin{itemize}[leftmargin=2em]
    \item[$L_0$] \textbf{Stable deterministic.} Fixed points or limit cycles, where the present state strongly constrains the future, e.g.,\ deterministic flows with non-positive largest Lyapunov exponent $\lambda_{\max} \le 0$.
    \item[$L_1$] \textbf{Weakly chaotic.}Deterministic dynamics with small positive instability and a longer but finite predictability horizon, $0 < \lambda_{\max}\tau \ll 1$.
    \item[$L_2$] \textbf{Strongly chaotic.} Deterministic dynamics with stronger instability and rapid trajectory divergence, $\lambda_{\max}\tau \gg 0$.
    \item[$L_3$] \textbf{Structured stochastic.} Processes that preserve selected linear statistics of a deterministic signal while disrupting nonlinear structure, e.g.,\ surrogate transformations $x_t^{(3)} = \mathcal{S}(x_t^{(2)})$ that preserve the power spectrum but destroy deterministic phase-space geometry.
    \item[$L_4$] \textbf{Unstructured stochastic.}\ i.i.d.\ noise with no predictive structure beyond the marginal distribution, $x_t \sim \mathcal{P}$ and $I(x_t ; x_{t+\tau}) = 0$ for $\tau>0$.
\end{itemize}
\end{definition}

This ladder reflects two axes of limited predictability. Along $L_0$--$L_2$, regimes remain deterministic, with predictability degrading as Lyapunov instability shortens the forecast horizon. The $L_2\!\to\!L_3$ transition instead destroys deterministic geometry: surrogate procedures preserve linear statistics while removing nonlinear flow structure. Level $L_4$ removes all temporal structure beyond the marginal distribution. Finer granularity can be introduced within adjacent levels by progressively increasing instability ($L_0$--$L_2$), disrupting deterministic geometry ($L_2$--$L_3$), or removing residual temporal dependence ($L_3$--$L_4$). This ordering parallels predictability-horizon arguments in nonlinear dynamical systems, where increasing trajectory divergence reduces forecastability.

\begin{figure}[t]
\centering
\includegraphics[width=\linewidth]{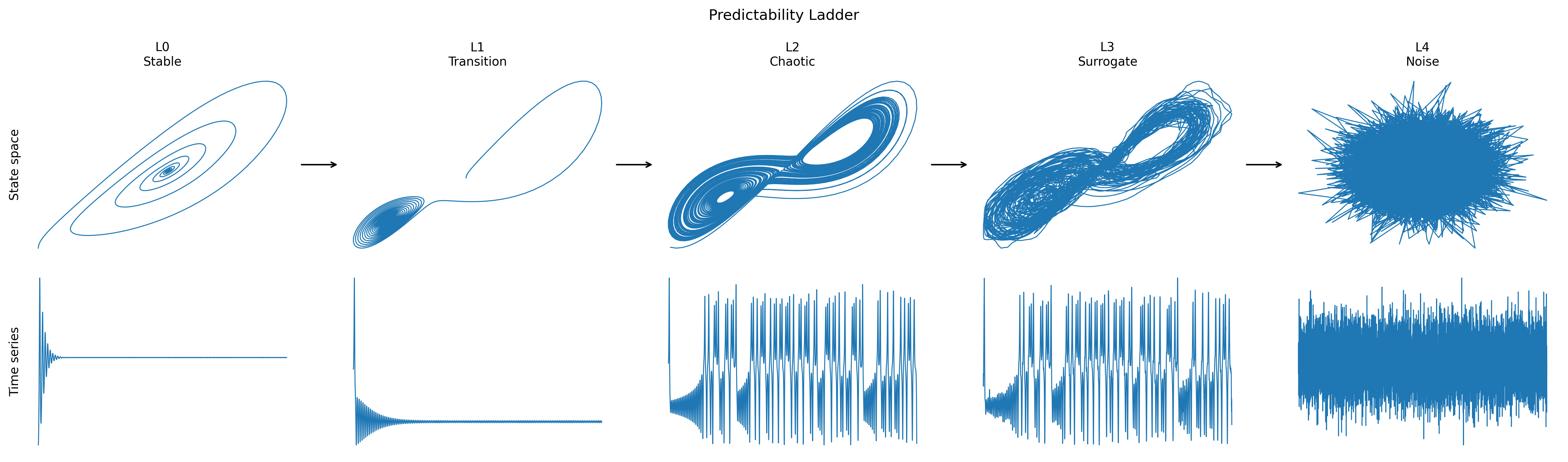}
\caption{
Predictability ladder construction. Top: state-space projections. Bottom: corresponding time series. Levels $L_0$--$L_2$ arise from the same dynamical system under different regimes. Level $L_3$ is obtained by surrogate generation that preserves selected linear statistics while disrupting deterministic structure. Level $L_4$ corresponds to unstructured noise.
}
\label{fig:ladder}
\end{figure}

\subsection{Ordinal Detectability}
\label{subsec:detectability}
\begin{definition}[Ordinal Detectability]
\label{def:detectability}
A scoring function $E_\theta:\mathbb{R}^{d \times T}\rightarrow\mathbb{R}$ is \emph{ordinally detectable} under a distribution $\mathcal{D}$ over
$(x,y)\in\mathcal{X}\times\{0,1,2,3,4\}$ if
\[
\mathbb{E}[E_\theta(x)\mid y=i] < \mathbb{E}[E_\theta(x)\mid y=j]
\qquad \forall\, i<j.
\]
\end{definition}
Let $\mu_k = \mathbb{E}[E_\theta(x)\mid y=k]$. To quantify global ordinal
structure beyond pairwise separation, we use the \emph{monotonicity slope}
\begin{equation}
\beta =
\frac{\sum_k (k-\bar{k})(\mu_k-\bar{\mu})}
     {\sum_k (k-\bar{k})^2},
\end{equation}
the least-squares slope of level means on ladder indices. A method may achieve a strong AUROC at a single boundary while producing $\beta \approx 0$ if the full ladder is not coherently ordered. We report $\beta$ both in-distribution and under transfer to systems withheld entirely from training.

\subsection{Cross-System Score Ambiguity}
\label{subsec:identifiability}

A scalar ordinal score is useful across systems only if its numerical values retain a stable interpretation; ranking alone does not guarantee this. Given a score $E:\mathcal{X}\to\mathbb{R}$ and thresholds $\tau_0 < \tau_1 < \tau_2 < \tau_3$, the induced predictor is  $h_E(x) = \sum_{m=0}^{3}\mathbf{1}\{E(x)>\tau_m\}$,  mapping $x$ to one of five ordered levels.

\begin{proposition}[Monotone reparameterization invariance]
\label{prop:monotone}
Let $f:\mathbb{R}\to\mathbb{R}$ be strictly increasing. The predictor induced by $E$ is unchanged under $\tilde{E} = f \circ E$ when thresholds are transformed to $\tilde{\tau}_m = f(\tau_m)$.
\emph{(Proof in Appendix~\ref{app:proof_monotone}.)}
\end{proposition}

Ordinal supervision therefore determines score \emph{ordering} but not its numerical \emph{coordinate}: the score is identifiable only up to a monotone reparameterization. While this ambiguity is harmless within a single system, it becomes consequential under transfer, where identical score values may correspond to different predictability regimes. Cross-system comparison thus requires a shared coordinate convention, which the anchor objective provides by fixing level-wise score locations globally.

\section{Method}
\label{sec:method}
GON has three components: 2-jet preprocessing, a temporal convolutional encoder, and a gauge-fixed ordinal objective.

\subsection{2-Jet Preprocessing}
\label{subsec:2jet}

Raw state coordinates carry system-specific scaling, orientation, and units. Representing each window by its 2-jet exposes local trajectory geometry independent of these system-specific factors. For a smooth trajectory $x(t)$, the 2-jet is the tuple
\[
(x(t),\dot{x}(t),\ddot{x}(t)),
\]
consisting of position, velocity, and acceleration. These quantities summarize local behavior up to second order and provide direct access to geometric features such as tangent direction and curvature. For trajectories generated by smooth ODEs, these components are linked by the governing flow. Surrogate procedures, by contrast, preserve selected linear statistics while disrupting the original nonlinear relation among position, velocity, and acceleration. This makes the 2-jet a natural representation for distinguishing deterministic trajectories from structured stochastic surrogates.

\paragraph{Implementation.}
Given a window $x\in\mathbb{R}^{d\times T}$, we first apply a Savitzky--Golay filter (window length $7$, polynomial order $3$; clamped to signal length and kept odd if shorter) to obtain a smoothed trajectory $\hat{x}$. Central finite differences are then used to estimate derivatives, as they provide symmetric second-order 
accurate approximations while minimizing phase bias in the local trajectory geometry:
\begin{align}
\dot{x}_t &= \frac{\hat{x}_{t+1}-\hat{x}_{t-1}}{2\Delta t}, \\
\ddot{x}_t &= \frac{\hat{x}_{t+1}-2\hat{x}_t+\hat{x}_{t-1}}{\Delta t^2}.
\end{align}
One timestep is removed from each boundary to avoid edge effects. The resulting representation is
\[
J^2(x) = [\hat{x},\,\dot{x},\,\ddot{x}]
\in \mathbb{R}^{3d \times (T-2)},
\]
which serves as the network input.

\subsection{Architecture}
\label{subsec:architecture}
The encoder is a temporal convolutional network (TCN) of dilated residual blocks
that maps the 2-jet input to a hidden sequence $h \in \mathbb{R}^{H\times T'}$.
After normalization, multi-scale average pooling aggregates $h$ into a vector
$z \in \mathbb{R}^{4H}$, which a two-layer MLP maps to a scalar score:
\begin{equation}
E_\theta(x) = \mathbf{W}_2\,\mathrm{GELU}\!\left(\mathrm{LN}(\mathbf{W}_1 z + b_1)\right) + b_2.
\end{equation}
Outputs are smoothly clipped via $E_\theta \leftarrow c\,\tanh(E_\theta/c)$, $c=5$, to keep scores aligned with the anchor range. Full details appear in Appendix~\ref{app:architecture}.

\subsection{Gauge-Fixed Ordinal Objective}
\label{subsec:objective}

Ranking-based supervision does not pin a unique score coordinate (Section~\ref{subsec:identifiability}). We fix one by assigning target locations
\[
\{t_k\}_{k=0}^4 = \{-4,-2,0,+2,+4\}
\]
and training the model so that each level concentrates around its anchor. These numeric targets are arbitrarily ordered anchors chosen to fix a shared coordinate; any monotonic rescaling yields an equivalent ordinal representation. Equal spacing is chosen for simplicity; the ordinal structure is invariant to any monotone rescaling of the targets, so the specific values affect only the absolute score scale and not relative ordering.
The objective is
\begin{equation}
\mathcal{L}(\theta)
= \lambda_a \sum_{k=0}^{4} (\mu_k - t_k)^2
+ \lambda_v \sum_{k=0}^{4} [\sigma_k - \sigma^*]_+^2,
\label{eq:loss}
\end{equation}
where $\mu_k$ and $\sigma_k$ are the mini-batch mean and standard deviation of scores for level~$k$, $[\cdot]_+=\max(0,\cdot)$, and $\sigma^*=0.3$. The anchor term fixes level-wise score locations, selecting a canonical representative from the class of monotone-equivalent solutions  (Appendix~\ref{app:gauge}); the variance term limits within-level dispersion,  together defining a shared affine coordinate across systems.

\paragraph{Relation to margin-based objectives.}
A natural alternative is margin loss $\mathcal{L}_{\text{margin}} =  \sum_{k=0}^{3}\max(0,\,\gamma-(\mu_{k+1}-\mu_k))$, which enforces adjacent-level separation but leaves the absolute score axis unfixed. Scores therefore remain non-comparable across systems even when levels are well-separated; Section~\ref{sec:ablation} tests this directly.

\subsection{Training Details}
\label{sec:training}
We set $\lambda_a=1.0$ with linear warmup over the first five epochs and $\lambda_v=0.01$. Optimization uses AdamW with learning rate $2\times10^{-3}$, weight decay $10^{-4}$, cosine decay over $30$ epochs, gradient clipping at $5.0$, and EMA decay $0.995$. Training uses a batch size of $64$. Weighted sampling balances ladder levels. The full experimental suite required approximately 40--50 GPU-hours 
on a single A100.

\paragraph{Data Augmentation.}
Augmentation is applied to raw trajectory windows $x \in \mathbb{R}^{d \times T}$ before 2-jet preprocessing and includes additive Gaussian noise ($\sigma_{\text{rel}}=0.05$), multiplicative scaling ($\times[0.9,1.1]$), temporal masking ($5\%$), linear drift ($\pm 0.05$), and causal time shifts ($\leq 5$ steps). Each operation is applied independently with probability $0.5$; the full pipeline 
activates with probability $0.7$. Augmentation precedes smoothing, so the filter attenuates discontinuities introduced by masking or time shifts before finite differences are computed.

\section{Experiments}
\label{sec:experiments}
The experiments address four questions:
\begin{enumerate}[leftmargin=1.5em, itemsep=2pt]
    \item Does the gauge-fixed coordinate remain calibrated on the source distribution, and does it transfer in zero-shot to held-out systems?
    \item Does pretraining provide a better initialization than scratch for adapting to unseen systems under limited labeled windows?
    \item Which objective component is responsible for cross-system coordinate stability?
    \item Does coherent 2-jet structure contribute beyond dimensionality-matched controls?
\end{enumerate}

\subsection{Experimental Setup}
\label{subsec:setup}

\paragraph{Source and target systems.}
Training uses 12 three-dimensional dissipative chaotic ODE systems spanning a range of attractor geometries: Chen, Chua, Duffing, Finance, Genesio--Tesi, Hastings--Powell, Lorenz-63, Lorenz-84, R\"ossler, Rucklidge, Shimizu--Morioka, and Halvorsen. Each system is simulated to produce trajectories of length $N=4096$, with system-specific timesteps chosen to resolve dominant dynamical timescales; $\Delta t$ is determined from the observed sampling rate and requires no system-level knowledge.

Ladder levels follow Definition~\ref{def:ladder}: deterministic regimes ($L_0$--$L_2$) are assigned using $\lambda_1$, $L_3$ is generated by applying surrogate procedures (AAFT, IAAFT, WLS, phase-shuffle) to $L_2$ trajectories, and $L_4$ is i.i.d.\ noise. Full simulation and labeling details appear in Appendix~\ref{app:data}.

Cross-system transfer is evaluated on five held-out systems: Thomas, Tigan, Newton--Leipnik, Rabinovich--Fabrikant, and the forced pendulum.

\paragraph{Windowing and preprocessing.}
Trajectories are segmented into windows of length $T=256$ with stride $128$. Inputs are normalized via per-trajectory, per-channel z-score standardization, requiring no source-distribution statistics and applying equally to any held-out system. After 2-jet preprocessing, each example is a $9\times 254$ tensor. Windows are the sole input unit; system-level information is unknown throughout. The adaptation experiment measures how many windows suffice to recalibrate a pretrained model to a new system under limited data.

\paragraph{Metrics.}
The monotonicity slope $\beta$ (Definition~\ref{def:detectability}) measures global ordinal coherence. Cross-method zero-shot comparisons use the scale-normalized $\beta_{\text{norm}}$; raw $\beta$ is used within the GON family, where the score scale is fixed by construction. Adjacent-pair AUROC at each boundary $(L_0,L_1)$ through $(L_3,L_4)$ serves as a diagnostic, since a method can separate one transition well while failing global coherence. For in-distribution calibration, we also report anchor drift
\[
\delta_{\text{anchor}}=\frac{1}{5}\sum_{k=0}^4 |\mu_k-t_k|,
\]
measuring how closely the learned level means match the anchor targets.

\subsection{Zero-Shot Evaluation}
\label{subsec:zeroshot}
Models are trained on the source systems and applied directly to the held-out systems.

\paragraph{In-distribution coordinate calibration.}
Before assessing transfer, it is useful to verify that the gauge-fixed score remains well aligned with its intended coordinate system on held-out source-domain data. Table~\ref{tab:calibration} reports level-wise score statistics on the test split of the training systems. GON closely matches the anchor targets, with mean anchor drift $\delta_{\text{anchor}}=0.0097$. The within-level standard deviations ($0.08$--$0.32$) remain well below the 2-unit spacing between adjacent anchors. 

\vspace{-6pt}
\begin{table}[ht]
\centering
\caption{In-distribution level calibration (mean $\pm$ std). Anchor targets 
$\{-4,-2,0,+2,+4\}$; mean anchor drift $\delta_\text{anchor}=0.0097$. Within-level standard deviations remain well below the 2-unit inter-anchor spacing, confirming well-separated source-distribution coordinates and supporting zero-shot transfer.}
\label{tab:calibration}
\small
\begin{tabular}{lrrrrr}
\toprule
 & $L_0$ & $L_1$ & $L_2$ & $L_3$ & $L_4$ \\
\midrule
Target
  & $-4.0$ & $-2.0$ & $0.0$ & $+2.0$ & $+4.0$ \\
GON
  & $-3.99{\pm}0.32$ & $-1.99{\pm}0.21$ & $-0.02{\pm}0.11$
  & $+2.00{\pm}0.23$ & $+4.01{\pm}0.08$ \\
\bottomrule
\end{tabular}
\end{table}

\begin{figure}[t]
\centering
\includegraphics[width=0.45\linewidth]{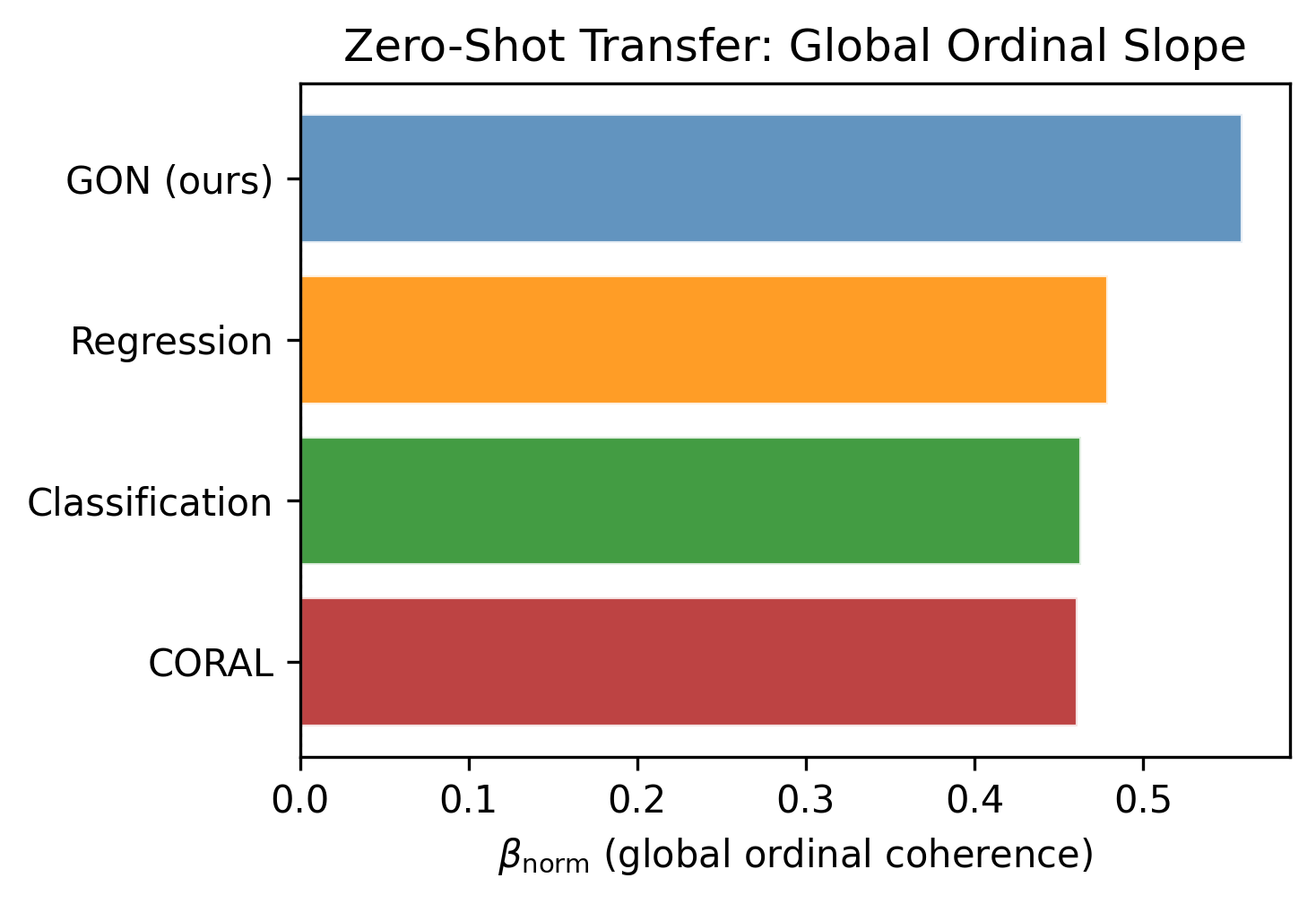}
\caption{Scale-normalized monotonicity slope $\beta_\text{norm}$ (raw $\beta$ divided by the inter-level step size of $2.0$, so $\beta_\text{norm}=1$ indicates perfectly monotone level-wise means) on five held-out systems under zero-shot transfer; higher is better. All supervised baselines cluster at  $\beta_\text{norm}\approx0.46$--$0.48$ regardless of training objective; GON achieves $0.559$ via a fixed shared ordinal coordinate.}
\label{fig:ordinal_Transfer}
\end{figure}

\paragraph{Cross-system zero-shot transfer.}
Figure~\ref{fig:ordinal_Transfer} compares GON with neural baselines on the held-out systems. All supervised baselines cluster tightly at $\beta_{\text{norm}} \approx 0.46$--$0.48$ despite using different objectives. GON achieves the highest value, $\beta_{\text{norm}}=0.559$. Full per-boundary AUROC and per-system neural baseline results are reported in Appendix~\ref{app:neural}. The transfer pattern is selective rather than uniform. The strongest zero-shot separation occurs at the stochastic boundary ($L_2 \to L_3$ and $L_3 \to L_4$), while deterministic-side boundaries remain weaker: $\auroc{01}$ is near chance and $\auroc{12}$ is only moderate (Appendix~\ref{app:neural}). The gauge-fixed score thus retains meaningful ordinal structure under distribution shift, with transfer concentrated where surrogate procedures most strongly disrupt nonlinear geometry.

\subsection{Adaptation from Pretrained Initialization}
\label{subsec:adaptation}
\begin{figure}[t]
    \centering
    \includegraphics[width=\linewidth, height=0.20\textheight, keepaspectratio=false]{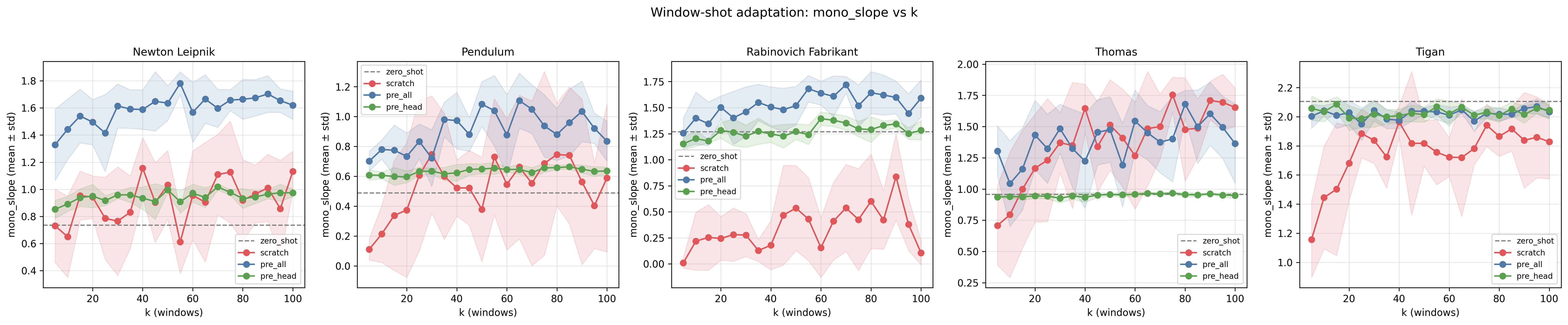}
    \caption{Adaptation from pretrained initialization vs.\ scratch per held-out system ($k \leq 100$ labeled windows, mean $\pm$ std over 5 seeds). Strategies: 
    \textbf{scratch} (random initialization), \textbf{pre\_head} (frozen encoder, 
    readout fine-tuned), \textbf{pre\_all} (full fine-tune from pretrained checkpoint).  Dashed line marks zero-shot GON performance. \textbf{pre\_all} consistently exceeds zero-shot with lower variance; \textbf{scratch} never recovers ordinal structure on  Rabinovich--Fabrikant even at $k=100$.}
    \label{fig:winshot}
\end{figure}

We evaluate whether GON pretraining provides a better initialization than training from scratch across $k \in \{5,10,\ldots,100\}$ labeled windows per held-out system, averaged over five random seeds. Three strategies are compared: \textbf{scratch} (random initialization), \textbf{pre\_head} (frozen encoder, readout fine-tuned), and \textbf{pre\_all} (full model fine-tuned from the pretrained checkpoint). Table~\ref{tab:adaptation} reports averages over all five systems; Figure~\ref{fig:winshot} and Appendix~\ref{app:adaptation} show per-system results and fine-tuning details.

\paragraph{Pretrained initialization is faster and less variable than scratch.}
At $k=5$, \textbf{pre\_all} leads scratch by $+0.776$ in $\beta$ ($1.319$ vs.\ $0.543$) and $+0.191$ in $\auroc{23}$ ($0.758$ vs.\ $0.567$), and this gap does not close at $k=100$ (Table~\ref{tab:adaptation}), indicating that the pretrained encoder encodes geometric structure that random initialization cannot recover within this window budget. \textbf{pre\_head} reliably exceeds scratch on $\auroc{23}$ but captures little of \textbf{pre\_all}'s gain in $\beta$, confirming that effective adaptation requires updating the encoder.

\vspace{-6pt}
\begin{table}[ht]
\centering
\caption{Adaptation: mean $\auroc{23}$ and $\beta$ averaged over 5 held-out systems and 5 random seeds (std across systems). \textbf{scratch}: random  initialization; \textbf{pre\_head}: frozen encoder, readout fine-tuned; \textbf{pre\_all}: full fine-tune from pretrained checkpoint. The  \textbf{pre\_all} advantage over scratch does not close at $k=100$.}
\label{tab:adaptation}
\small
\begin{tabular}{l rrr rrr}
\toprule
 & \multicolumn{3}{c}{$\auroc{23}$}
 & \multicolumn{3}{c}{$\beta$} \\
\cmidrule(lr){2-4}\cmidrule(lr){5-7}
$k$ & scratch & pre\_head & pre\_all
    & scratch & pre\_head & pre\_all \\
\midrule
$5$   & $0.567$\sd{0.14} & $0.717$\sd{0.25} & $\mathbf{0.758}$\sd{0.21}
      & $0.543$\sd{0.48} & $1.122$\sd{0.56} & $\mathbf{1.319}$\sd{0.46} \\
$10$  & $0.566$\sd{0.13} & $0.714$\sd{0.25} & $\mathbf{0.762}$\sd{0.21}
      & $0.664$\sd{0.51} & $1.135$\sd{0.55} & $\mathbf{1.341}$\sd{0.48} \\
$25$  & $0.567$\sd{0.13} & $0.718$\sd{0.25} & $\mathbf{0.765}$\sd{0.19}
      & $0.958$\sd{0.62} & $1.149$\sd{0.52} & $\mathbf{1.385}$\sd{0.40} \\
$50$  & $0.580$\sd{0.14} & $0.723$\sd{0.26} & $\mathbf{0.795}$\sd{0.17}
      & $1.055$\sd{0.62} & $1.177$\sd{0.52} & $\mathbf{1.551}$\sd{0.34} \\
$100$ & $0.562$\sd{0.15} & $0.723$\sd{0.25} & $\mathbf{0.782}$\sd{0.18}
      & $1.062$\sd{0.72} & $1.177$\sd{0.54} & $\mathbf{1.489}$\sd{0.44} \\
\bottomrule
\end{tabular}
\end{table}

\paragraph{Transfer depth scales with geometric proximity.}
Systems with polynomial or exponential vector fields represented in the source distribution (Rabinovich--Fabrikant, Newton--Leipnik, Tigan) adapt immediately or are solved at zero-shot. Rabinovich--Fabrikant is the clearest case: \textbf{scratch} yields $\beta\approx0.01$--$0.54$ across all $k$, never recovering ordinal structure, whereas \textbf{pre\_all} exceeds zero-shot from $k=10$. Thomas and the forced pendulum are harder: both contain state-dependent sinusoidal vector-field terms absent from training (e.g.,\ $\sin\theta$, $\sin y$), producing unseen oscillatory curvature in the 2-jet. In both cases, \textbf{pre\_all} recovers global ordinal structure but fails to resolve the $L_2\to L_3$ boundary, identifying sinusoidal vector-field geometry as the primary gap in source-distribution coverage.

\subsection{Ablations}
\label{sec:ablation}

The ablations ask which parts of the method matter for transfer.

\paragraph{Objective ablation: why the anchor matters.}
Table~\ref{tab:loss_ablation} shows that coordinate stability, not ordinal coherence alone, is critical for transfer. Replacing the anchor with a margin objective leaves in-distribution AUROC unchanged by at most $0.010$ across adjacent pairs and produces comparable scale-normalized transfer ($\beta_\text{norm}=0.554$ vs.\ $0.559$), yet anchor drift rises from $0.005$ to $2.556$. A model can therefore retain ordinal coherence while losing coordinate stability, confirming that relative level separation 
is insufficient for cross-system comparability: without a fixed coordinate, scores on unseen systems carry no consistent numerical meaning. The variance 
term plays a secondary but consistent role: removing it increases within-level spread from $0.195$ to $0.230$ and lowers transfer $\beta$ from $1.118$ to $1.055$.
\vspace{-6pt}
\begin{table}[ht]
\centering
\caption{
Loss ablation. Anchor drift measures coordinate stability; transfer $\beta$ 
measures ordinal transfer on held-out systems. Anchor-based objectives yield 
stable coordinates and better transfer.
}
\label{tab:loss_ablation}
\small
\begin{tabular}{lrrrr}
\toprule
Objective
  & $\delta_\text{anchor}$ (indist)
  & $\beta$ (indist)
  & $\beta$ (Transfer)
  & $\bar{\sigma}$ (indist) \\
\midrule
Anchor $+$ Variance (ours)
  & \textbf{0.005} & 2.002 & \textbf{1.118} & \textbf{0.195} \\
Margin $+$ Variance
  & 2.556          & 1.075 & 0.596          & 0.244          \\
Anchor only
  & 0.014          & 2.006 & 1.055          & 0.230          \\
\bottomrule
\end{tabular}
\end{table}

\paragraph{Input ablation: why the 2-jet helps.}
Table~\ref{tab:jet_ablation} shows that coherent derivative structure improves transfer beyond raw states. Adding first derivatives improves $\beta$ from $1.132$ to $1.336$; the full 2-jet improves further to $\beta=1.504$ and $\auroc{23}=0.846$. Shuffling derivatives across time drops $\beta$ to $0.983$, confirming that temporal coherence among position, velocity, and acceleration drives the gain. A control replacing derivatives with per-pass Gaussian noise remains near the raw-state baseline, ruling out channel count.
\vspace{-6pt}
\begin{table}[ht]
\centering
\caption{
Jet ablation: transfer $\beta$ and $\auroc{23}$ over five held-out systems
(anchor$+$variance, no Savitzky--Golay smoothing). Coherent 2-jet features
improve transfer; shuffled and random ($\dagger$) controls degrade performance.
}
\label{tab:jet_ablation}
\small
\begin{tabular}{lrr}
\toprule
Input representation & Transfer $\beta$ & Transfer $\auroc{23}$ \\
\midrule
Raw state ($x$ only)                          & 1.132 & 0.791 \\
1-jet ($x$, $\dot{x}$)                        & 1.336 & 0.789 \\
\textbf{2-jet} ($x$, $\dot{x}$, $\ddot{x}$)  & \textbf{1.504} & \textbf{0.846} \\
\midrule
Shuffled 2-jet$^\dagger$ (temporal control)   & 0.983 & 0.737 \\
Random channels$^\dagger$ (dim.\ control)     & 1.174 & 0.734 \\
\bottomrule
\end{tabular}
\end{table}

\paragraph{Noise robustness.}
GON remains robust under moderate corruption, with $\auroc{23}$ staying 
above $0.89$ at ${\geq}\,20$\,dB SNR and $30\%$ temporal dropout. Performance degrades gracefully in $\beta$, indicating reduced transfer strength before loss of discriminability. Full results appear in Appendix~\ref{app:noise}.

\section{Discussion}
\label{sec:discussion}

\paragraph{A fixed score coordinate enables transferable predictability estimates.}
Replacing the anchor term with a margin loss leaves in-distribution discrimination nearly unchanged but increases anchor drift and reduces transfer $\beta$ by $47\%$ (Table~\ref{tab:loss_ablation}). Source-domain separability alone, therefore, does not ensure that a scalar score retains a stable meaning on unseen systems. Gauge fixing instead learns a shared ordinal coordinate, enabling predictability comparisons that are transferred across heterogeneous dynamics.

\paragraph{Transfer concentrates where predictability is structurally limited.}
Zero-shot transfer is strongest at the stochastic boundary ($L_2 \to L_3$ 
and $L_3 \to L_4$); deterministic-side boundaries remain weaker. The same 
asymmetry holds under perturbation: $\auroc{23}$ is more robust than 
$\auroc{01}$ to noise and dropout. GON captures geometric structure disrupted 
by surrogate generation, whereas finer distinctions within deterministic chaos 
are more system-specific.

\paragraph{Geometric representations support cross-system generalization.}
The 2-jet representation improves transfer by exposing local trajectory geometry. The benefit depends on coherent relationships among position, velocity, and acceleration; disrupting this alignment reduces the transfer even at a matched dimensionality. This supports the view that transferable predictability cues arise from shared geometric structure rather than system-specific statistics.

\paragraph{Pretraining helps, but transfer depth is system-dependent.}
Systems whose vector fields resemble those seen in training adapt immediately or are solved at zero-shot. In contrast, systems with unseen structure (e.g., sinusoidal forcing in Thomas and the forced pendulum) require more data and may fail to resolve finer boundaries. Performance degrades smoothly with geometric distance from the source distribution.

\paragraph{Scope, limitations, and broader implications.}
The benchmark covers 12 source and 5 target systems, all low-dimensional 
and dissipative; extensions to higher-dimensional systems, conservative 
dynamics, or real-world signals, remain unestablished. 
This work takes one of the earliest steps toward a general theory of 
transferable predictability estimation, establishing gauge-fixing as a 
necessary condition for cross-system ordinal comparability. Predictive 
structure in real systems is often limited by chaos, stochastic forcing, 
or their interaction, arising across weather and climate dynamics, turbulent 
flows, biological oscillations, financial processes, and engineered systems. 
By learning a transferable ordinal predictability coordinate, the framework 
enables cross-system comparison without system-specific calibration, 
informing model selection, early-warning diagnostics, and the design of 
resilient coupled human--natural systems. The method operates exclusively 
on synthetic ODE signals, involves no personal data, and has no foreseeable 
negative societal impacts.

\section{Conclusion}
This paper identifies gauge-fixing as the missing ingredient for transferable ordinal predictability scores. The central challenge is not ranking regimes 
within a single system, but assigning scores whose numerical values retain a 
consistent interpretation across systems. The loss ablation makes the point 
cleanly: replacing the anchor with a margin objective leaves the source-domain 
discrimination nearly unchanged while reducing transfer $\beta$ by $47\%$. 
Ordinal coherence is robust to 4-bit quantization and Gaussian noise down to 
20\,dB SNR, confirming the learned coordinate survives common observational 
degradations. Pairwise discrimination and globally coherent ordinal scoring 
are distinct properties that can dissociate, and a stable coordinate is 
necessary to bridge them under transfer. The benchmark is limited to 
low-dimensional dissipative systems; this bound does not invalidate the core 
finding, since the ablation isolates coordinate convention as the operative 
variable within a fixed system class. Extension to higher-dimensional 
attractors, partial observability, and real-world signals in weather, climate, 
and physiology remain the most consequential direction for future work.
\bibliographystyle{plainnat}
\bibliography{references}

@book{agresti2010analysis,
  title={Analysis of ordinal categorical data},
  author={Agresti, Alan},
  year={2010},
  publisher={John Wiley \& Sons}
}

@article{cao2020rank,
  title={Rank consistent ordinal regression for neural networks with application to age estimation},
  author={Cao, Wenzhi and Mirjalili, Vahid and Raschka, Sebastian},
  journal={Pattern Recognition Letters},
  volume={140},
  pages={325--331},
  year={2020},
  publisher={Elsevier}
}

@article{zha2023rank,
  title={Rank-n-contrast: Learning continuous representations for regression},
  author={Zha, Kaiwen and Cao, Peng and Son, Jeany and Yang, Yuzhe and Katabi, Dina},
  journal={Advances in Neural Information Processing Systems},
  volume={36},
  pages={17882--17903},
  year={2023}
}

@article{wolf1985determining,
  title={Determining Lyapunov exponents from a time series},
  author={Wolf, Alan and Swift, Jack B and Swinney, Harry L and Vastano, John A},
  journal={Physica D: nonlinear phenomena},
  volume={16},
  number={3},
  pages={285--317},
  year={1985},
  publisher={Elsevier}
}

@article{rosenstein1993practical,
  title={A practical method for calculating largest Lyapunov exponents from small data sets},
  author={Rosenstein, Michael T and Collins, James J and De Luca, Carlo J},
  journal={Physica D: Nonlinear Phenomena},
  volume={65},
  number={1-2},
  pages={117--134},
  year={1993},
  publisher={Elsevier}
}

@article{theiler1992testing,
  title={Testing for nonlinearity in time series: the method of surrogate data},
  author={Theiler, James and Eubank, Stephen and Longtin, Andr{\'e} and Galdrikian, Bryan and Farmer, J Doyne},
  journal={Physica D: Nonlinear Phenomena},
  volume={58},
  number={1-4},
  pages={77--94},
  year={1992},
  publisher={Elsevier}
}

@article{schreiber1996improved,
  title={Improved surrogate data for nonlinearity tests},
  author={Schreiber, Thomas and Schmitz, Andreas},
  journal={Physical review letters},
  volume={77},
  number={4},
  pages={635},
  year={1996},
  publisher={APS}
}

@article{keylock2006constrained,
  title={Constrained surrogate time series with preservation of the mean and variance structure},
  author={Keylock, CJ},
  journal={Physical Review E—Statistical, Nonlinear, and Soft Matter Physics},
  volume={73},
  number={3},
  pages={036707},
  year={2006},
  publisher={APS}
}

@article{chen2018neural,
  title={Neural ordinary differential equations},
  author={Chen, Ricky TQ and Rubanova, Yulia and Bettencourt, Jesse and Duvenaud, David K},
  journal={Advances in neural information processing systems},
  volume={31},
  year={2018}
}

@article{cranmer2020lagrangian,
  title={Lagrangian neural networks},
  author={Cranmer, Miles and Greydanus, Sam and Hoyer, Stephan and Battaglia, Peter and Spergel, David and Ho, Shirley},
  journal={arXiv preprint arXiv:2003.04630},
  year={2020}
}

@inproceedings{finn2017model,
  title={Model-agnostic meta-learning for fast adaptation of deep networks},
  author={Finn, Chelsea and Abbeel, Pieter and Levine, Sergey},
  booktitle={International conference on machine learning},
  pages={1126--1135},
  year={2017},
  organization={PMLR}
}

@inproceedings{kirchmeyer2022generalizing,
  title={Generalizing to new physical systems via context-informed dynamics model},
  author={Kirchmeyer, Matthieu and Yin, Yuan and Don{\`a}, J{\'e}r{\'e}mie and Baskiotis, Nicolas and Rakotomamonjy, Alain and Gallinari, Patrick},
  booktitle={International conference on machine learning},
  pages={11283--11301},
  year={2022},
  organization={PMLR}
}

@article{li2020fourier,
  title={Fourier neural operator for parametric partial differential equations},
  author={Li, Zongyi and Kovachki, Nikola and Azizzadenesheli, Kamyar and Liu, Burigede and Bhattacharya, Kaushik and Stuart, Andrew and Anandkumar, Anima},
  journal={arXiv preprint arXiv:2010.08895},
  year={2020}
}

@article{lu2021learning,
  title={Learning nonlinear operators via DeepONet based on the universal approximation theorem of operators},
  author={Lu, Lu and Jin, Pengzhan and Pang, Guofei and Zhang, Zhongqiang and Karniadakis, George Em},
  journal={Nature machine intelligence},
  volume={3},
  number={3},
  pages={218--229},
  year={2021},
  publisher={Nature Publishing Group UK London}
}

@article{lai2025panda,
  title={Panda: A pretrained forecast model for chaotic dynamics},
  author={Lai, Jeffrey and Bao, Anthony and Gilpin, William},
  journal={arXiv preprint arXiv:2505.13755},
  year={2025}
}

@article{pathak2018model,
  title={Model-free prediction of large spatiotemporally chaotic systems from data: A reservoir computing approach},
  author={Pathak, Jaideep and Hunt, Brian and Girvan, Michelle and Lu, Zhixin and Ott, Edward},
  journal={Physical review letters},
  volume={120},
  number={2},
  pages={024102},
  year={2018},
  publisher={APS}
}

@article{bury2020detecting,
  title={Detecting and distinguishing tipping points using spectral early warning signals},
  author={Bury, Thomas M and Bauch, Chris T and Anand, Madhur},
  journal={Journal of the Royal Society Interface},
  volume={17},
  number={170},
  year={2020},
  publisher={The Royal Society}
}

@article{chen1999yet,
  title={Yet another chaotic attractor},
  author={Chen, Guanrong and Ueta, Tetsushi},
  journal={International Journal of Bifurcation and chaos},
  volume={9},
  number={07},
  pages={1465--1466},
  year={1999},
  publisher={World Scientific}
}

@article{chua1986double,
  title={The double scroll family},
  author={Chua, LEONO and Komuro, Motomasa and Matsumoto, Takashi},
  journal={IEEE transactions on circuits and systems},
  volume={33},
  number={11},
  pages={1072--1118},
  year={1986},
  publisher={IEEE}
}

@book{duffing1918erzwungene,
  title     = {Erzwungene Schwingungen bei ver{\"a}nderlicher Eigenfrequenz und ihre technische Bedeutung},
  author    = {Duffing, Georg},
  series    = {Sammlung Vieweg},
  number    = {41-42},
  year      = {1918},
  publisher = {Vieweg}
}

@article{hastings1991chaos,
  title={Chaos in a three-species food chain},
  author={Hastings, Alan and Powell, Thomas},
  journal={Ecology},
  volume={72},
  number={3},
  pages={896--903},
  year={1991},
  publisher={Wiley Online Library}
}

@article{lorenz1963deterministic,
  author  = {Lorenz, Edward N.},
  title   = {Deterministic nonperiodic flow},
  journal = {Journal of the Atmospheric Sciences},
  volume  = {20},
  number  = {2},
  pages   = {130--141},
  year    = {1963},
}

@article{rossler1976equation,
  title={An equation for continuous chaos},
  author={R{\"o}ssler, Otto E},
  journal={Physics Letters A},
  volume={57},
  number={5},
  pages={397--398},
  year={1976},
  publisher={Elsevier}
}

@article{shimizu1980bifurcation,
  title={On the bifurcation of a symmetric limit cycle to an asymmetric one in a simple model},
  author={Shimizu, T and Morioka, N},
  journal={Physics Letters A},
  volume={76},
  number={3-4},
  pages={201--204},
  year={1980},
  publisher={Elsevier}
}

@article{cai2007new,
  title={A new finance chaotic attractor},
  author={Cai, Guoliang and Huang, Juanjuan},
  journal={International Journal of Nonlinear Science},
  volume={3},
  number={3},
  pages={213--220},
  year={2007}
}

@article{genesio1992harmonic,
  title={Harmonic balance methods for the analysis of chaotic dynamics in nonlinear systems},
  author={Genesio, Roberto and Tesi, Alberto},
  journal={Automatica},
  volume={28},
  number={3},
  pages={531--548},
  year={1992},
  publisher={Elsevier}
}

@article{lorenz1984irregularity,
  title={Irregularity: A fundamental property of the atmosphere},
  author={Lorenz, Edward N},
  journal={Tellus A},
  volume={36},
  number={2},
  pages={98--110},
  year={1984},
  publisher={Wiley Online Library}
}

@article{Rucklidge_1992, title={Chaos in models of double convection}, volume={237}, ISSN={1469-7645}, url={http://dx.doi.org/10.1017/s0022112092003392}, DOI={10.1017/s0022112092003392}, journal={Journal of Fluid Mechanics}, publisher={Cambridge University Press (CUP)}, author={Rucklidge, A. M.}, year={1992}, month=apr, pages={209–229} }

@article{thomas1999deterministic,
  author  = {Thomas, Ren{\'e}},
  title   = {Deterministic Chaos Seen in Terms of Feedback Circuits: Analysis, Synthesis, ``{L}abyrinth Chaos''},
  journal = {International Journal of Bifurcation and Chaos},
  volume  = {9},
  number  = {10},
  pages   = {1889--1905},
  year    = {1999},
  doi     = {10.1142/S0218127499001383}
}

@article{Tigan_2008, title={Analysis of a 3D chaotic system}, volume={36}, ISSN={0960-0779}, url={http://dx.doi.org/10.1016/j.chaos.2006.07.052}, DOI={10.1016/j.chaos.2006.07.052}, number={5}, journal={Chaos, Solitons \& Fractals}, publisher={Elsevier BV}, author={Tigan, Gheorghe and Opriş, Dumitru}, year={2008}, month=jun, pages={1315–1319} }

@article{Leipnik_1981, title={Double strange attractors in rigid body motion with linear feedback control}, volume={86}, ISSN={0375-9601}, url={http://dx.doi.org/10.1016/0375-9601(81)90165-1}, DOI={10.1016/0375-9601(81)90165-1}, number={2}, journal={Physics Letters A}, publisher={Elsevier BV}, author={Leipnik, R.B. and Newton, T.A.}, year={1981}, month=nov, pages={63–67} }

@article{d1982chaotic,
  title={Chaotic states and routes to chaos in the forced pendulum},
  author={d'Humieres, D and Beasley, MR and Huberman, BA and Libchaber, A},
  journal={Physical Review A},
  volume={26},
  number={6},
  pages={3483},
  year={1982},
  publisher={APS}
}

@article{bandt2002permutation,
  author  = {Bandt, Christoph and Pompe, Bernd},
  title   = {Permutation entropy: a natural complexity measure for time series},
  journal = {Physical Review Letters},
  volume  = {88},
  number  = {17},
  pages   = {174102},
  year    = {2002}
}

@article{richman2000physiological,
  author  = {Richman, Joshua S. and Moorman, J. Randall},
  title   = {Physiological time-series analysis using approximate entropy and sample entropy},
  journal = {American Journal of Physiology--Heart and Circulatory Physiology},
  volume  = {278},
  number  = {6},
  pages   = {H2039--H2049},
  year    = {2000}
}

@article{sugihara1990,
  author  = {Sugihara, George and May, Robert M.},
  title   = {Nonlinear Forecasting as a Way of Distinguishing Chaos from
             Measurement Error in Time Series},
  journal = {Nature},
  year    = {1990},
  volume  = {344},
  pages   = {734--741},
  doi     = {10.1038/344734a0}
}

@article{scheffer2009early,
  title     = {Early-warning signals for critical transitions},
  author    = {Scheffer, Marten and Bascompte, Jordi and Brock, William A. and
               Brovkin, Victor and Carpenter, Stephen R. and Dakos, Vasilis and
               H{\"o}hne, Hermann and van Nes, Egbert H. and Rietkerk, Max and
               Sugihara, George},
  journal   = {Nature},
  volume    = {461},
  number    = {7260},
  pages     = {53--59},
  year      = {2009},
  publisher = {Nature Publishing Group}
}

@article{kaplan1992direct,
  title     = {Direct test for determinism in a time series},
  author    = {Kaplan, Daniel T. and Glass, Leon},
  journal   = {Physical Review Letters},
  volume    = {68},
  number    = {4},
  pages     = {427--430},
  year      = {1992},
  publisher = {American Physical Society}
}

@article{rabinovich1979stochastic,
  title={Stochastic self-modulation of waves in nonequilibrium media},
  author={Rabinovich, Mikhail I and Fabrikant, Anatoly L},
  journal={J. Exp. Theor. Phys},
  volume={77},
  pages={617--629},
  year={1979}
}

@book{sprott2010elegant,
  title     = {Elegant Chaos: Algebraically Simple Chaotic Flows},
  author    = {Sprott, Julien Clinton},
  year      = {2010},
  publisher = {World Scientific},
  address   = {Singapore}
}

@article{koenderink1987representation,
  title={Representation of local geometry in the visual system},
  author={Koenderink, Jan J and van Doorn, Andrea J},
  journal={Biological cybernetics},
  volume={55},
  number={6},
  pages={367--375},
  year={1987},
  publisher={Springer}
}

@article{mccullagh1980regression,
  title={Regression models for ordinal data},
  author={McCullagh, Peter},
  journal={Journal of the Royal Statistical Society: Series B (Methodological)},
  volume={42},
  number={2},
  pages={109--127},
  year={1980},
  publisher={Wiley Online Library}
}

@inproceedings{kim2020proxy,
  title={Proxy anchor loss for deep metric learning},
  author={Kim, Sungyeon and Kim, Dongwon and Cho, Minsu and Kwak, Suha},
  booktitle={Proceedings of the IEEE/CVF conference on computer vision and pattern recognition},
  pages={3238--3247},
  year={2020}
}

@article{schreiber2000surrogate,
  title={Surrogate time series},
  author={Schreiber, Thomas and Schmitz, Andreas},
  journal={Physica D: Nonlinear Phenomena},
  volume={142},
  number={3-4},
  pages={346--382},
  year={2000},
  publisher={Elsevier}
}

\clearpage
\appendix
\section*{Appendix}
\addcontentsline{toc}{section}{Appendix}

\section{Related Work}
\label{app:related}

\paragraph{Determinism versus stochasticity.}
Geometric diagnostics~\citep{sugihara1990,kaplan1992direct} and prediction-based tests study rejection of the stochastic null hypothesis within a single system. Permutation entropy~\citep{bandt2002permutation} and sample entropy~\citep{richman2000physiological} provide scalar complexity measures widely used as binary indicators of dynamical complexity. The present work places this binary separation within a broader ordinal hierarchy, where pairwise discriminability at a single boundary and global ordinal coherence across levels need not coincide.

\paragraph{Chaos detection and surrogate testing.}
The classical toolbox for the binary problem is surrogate hypothesis testing~\citep{theiler1992testing,schreiber1996improved}, which constructs a null
distribution from the signal and tests whether its nonlinear statistics exceed
those of matched stochastic alternatives. Complementary approaches estimate the
maximal Lyapunov exponent directly~\citep{wolf1985determining,rosenstein1993practical}. Both are designed for single-system analysis: each new signal requires new surrogates or freshly tuned estimators, and both address a binary question. The present work considers an ordinal formulation in which predictability is represented on a continuum, with scores intended to remain comparable across systems without per-signal calibration.

\paragraph{Early warning signals.}
Critical slowing down theory~\citep{scheffer2009early} and spectral precursor
methods~\citep{bury2020detecting} study statistical signatures of approaching
bifurcations. The predictability ladder subsumes early-warning detection as a
special case: $L_1$ is precisely the transition regime where such signatures
emerge. The present benchmark does not directly evaluate early-warning performance. Transfer is strongest at the stochastic boundary rather than the deterministic-side transition, and calibrated cross-system early-warning detection remains a natural extension.

\paragraph{Ordinal learning.}
Cumulative link models~\citep{agresti2010analysis} and neural extensions such as CORAL~\citep{cao2020rank} enforce a consistent rank structure across ordered labels. Ordinal contrastive methods~\citep{zha2023rank} encode ordering in representation space. Proxy-based metric learning~\citep{kim2020proxy} uses class-level anchor prototypes to stabilize embedding geometry. These formulations study ordering within a single system or dataset, while the setting considered here involves scores that are intended to remain comparable across heterogeneous systems.

\paragraph{Physics-informed and geometric representations.}
Neural differential equation models~\citep{chen2018neural} and
Lagrangian networks~\citep{cranmer2020lagrangian} encode geometric and
physics-aware inductive biases in learned representations. Reservoir computing has been applied to chaotic
forecasting~\citep{pathak2018model}. The 2-jet representation used here is
grounded in differential geometry: position, velocity, and acceleration jointly
encode local trajectory curvature in a way that is invariant to system-specific
scaling and is disrupted by surrogate generation procedures.

\paragraph{Transfer and meta-learning for dynamical systems.}
Gradient-based meta-learning~\citep{finn2017model} has been applied to dynamical systems forecasting~\citep{kirchmeyer2022generalizing}. Neural operators~\citep{li2020fourier,lu2021learning} learn solution operators across parameter families. \citet{lai2025panda} trained a transformer on a large synthetic corpus of chaotic systems and reported zero-shot forecasting on unseen ODEs and PDEs. These approaches study predictive generalization across systems under supervised forecasting objectives. The setting considered here focuses on learning representations of predictability regimes, where the goal is to assign comparable ordinal scores rather than to predict future states.

\section{Proof of Proposition~\ref{prop:monotone}}
\label{app:proof_monotone}
Let $f:\mathbb{R}\to\mathbb{R}$ be strictly increasing. For any threshold $\tau_m$ and input $x$,
\[
E(x) > \tau_m \iff f(E(x)) > f(\tau_m),
\]
since strictly increasing functions preserve the direction of all inequalities. The induced ordinal predictor $h_E(x) = \sum_{m=0}^{3}\mathbf{1}\{E(x) > \tau_m\}$ depends only on these threshold comparisons. Under the transformed score $\tilde{E} = f \circ E$ with transformed thresholds $\tilde{\tau}_m = f(\tau_m)$, each comparison $\tilde{E}(x) > \tilde{\tau}_m$ is equivalent to $E(x) > \tau_m$. Therefore, $h_{\tilde{E}}(x) = h_E(x)$ for all $x$, and the induced ordinal prediction is unchanged. \qed

\section{Architecture Details}
\label{app:architecture}

\subsection{Temporal Convolutional Encoder}

The encoder is a temporal convolutional network (TCN) composed of $B = 6$ residual blocks with exponentially increasing dilation factors $[1, 2, 4, 8, 16, 32]$. Each block consists of two 1D dilated convolutions with kernel size $k = 3$, followed by Group Normalization and GELU activation. Residual connections are applied at the block level.

The hidden dimension is fixed at $H = 128$ across all layers.

\subsection{Receptive Field}

The effective receptive field of the network is
\begin{equation}
\mathcal{R}
=
1 + 2(k - 1)\sum_{i=0}^{B-1} 2^i
=
253,
\end{equation}
which covers nearly the full temporal extent of the input after 2-jet boundary trimming ($T' = T - 2$). This allows the model to capture long-range temporal dependencies relevant for distinguishing predictability regimes.

\subsection{Readout and Multi-Scale Aggregation}
The encoder produces a feature map $h \in \mathbb{R}^{H \times T'}$. RMS normalization is applied before aggregation. Average pooling is performed at scales $s \in \{4, 16, 64, 256\}$ to capture structure at multiple temporal resolutions; for $s > T'$ the kernel is clamped to $k = \min(s, T')$, so at scale $256$ this reduces to a global average over all $254$ timesteps. Pooling uses no padding; non-integer-multiple tails are truncated by \texttt{F.avg\_pool1d} with default \texttt{padding=0}. The pooled representations are concatenated to form a feature vector $z \in \mathbb{R}^{4H}$.

\subsection{MLP Head and Output Scaling}
The pooled feature vector $z \in \mathbb{R}^{4H}$ (dimension $512$) is 
mapped to a scalar score via a two-layer MLP with hidden dimension $H=128$:
\begin{equation}
E_\theta(x) = \mathbf{W}_2\,\mathrm{GELU}\!\left(\mathrm{LN}(\mathbf{W}_1 z + b_1)\right) + b_2,
\end{equation}
where $\mathbf{W}_1 \in \mathbb{R}^{128 \times 512}$ and $\mathbf{W}_2 \in \mathbb{R}^{1 \times 128}$. To prevent unbounded growth and maintain consistency with the fixed anchor targets, the output is smoothly clipped:
\[
E_\theta \leftarrow c \tanh(E_\theta / c), \quad c = 5.
\]
The full model contains $616{,}233$ trainable parameters.

\section{Data Generation and Labeling Details}
\label{app:data}

\subsection{Numerical Integration}

Each system is simulated to produce $N = 4096$ uniformly sampled time steps.  The timestep $\Delta t$ is selected to resolve the dominant dynamical timescale of each system.  We estimate a characteristic timescale $\tau$ from the mean inter-peak interval of the trajectory,  and set $\Delta t$ such that approximately 64 dominant cycles are covered, with $\Delta t$  clamped to $[10^{-4}, 10^{-1}]$.

Integration is performed using either a fixed-step fourth-order Runge–Kutta method or the adaptive Dormand–Prince (DOPRI5) solver, depending on system stiffness.

\subsection{Predictability Ladder Construction}
Deterministic regimes follow Definition~\ref{def:ladder}, with $\lambda_1$
estimated via~\citet{rosenstein1993practical}. $L_0$ corresponds to
$\lambda_1 < 0$; $L_1$ and $L_2$ are distinguished by the relative magnitude
of $\lambda_1$, cross-checked by visual inspection of attractor phase portraits.
Because attractor geometries differ substantially across systems, no single
numerical threshold is applied uniformly; regime boundaries are confirmed by
qualitative separation in both the exponent estimate and the phase portrait.
These boundaries are fixed at data generation time and held constant throughout
all experiments; the qualitative inspection step affects only this initial label
assignment and does not enter any learned component of the method.
$L_3$ is generated from $L_2$ trajectories using AAFT, IAAFT, WLS, and phase
randomization~\citep{theiler1992testing,schreiber1996improved,keylock2006constrained,
schreiber2000surrogate}. $L_4$ is i.i.d.\ Gaussian noise matched in length and
dimensionality.

\subsection{Data Splits}

Train/validation/test splits are performed at the trajectory level using an 80/10/10 partition.  All normalization statistics are computed on the training split and held fixed for validation and testing.

\section{Gauge Fixing and Ordinal Identifiability}
\label{app:gauge}

Ordinal regression determines only the ordering of latent scores. For any
strictly increasing $f$, the transformed score $\tilde{E}(x)=f(E(x))$ induces identical ordinal decisions under the corresponding threshold transformation, so $E$ is identifiable only up to a monotone reparameterization \citep{mccullagh1980regression}. Within a single system, this ambiguity is harmless: thresholds are learned jointly with the score, and ranking suffices. Under cross-system transfer, it becomes consequential because identical ordinal predictions on two systems need not assign the same numeric value to the same predictability regime.

The anchor objective removes this degree of freedom by fixing the level-wise score means,
\[
\mathbb{E}[E(x)\mid y=k] = t_k,
\]
standardizing the macroscopic location and scale of the score distribution, breaking the unbounded gauge freedom of ordinal regression. The variance term further restricts within-level dispersion. Together, they establish a shared coordinate convention across systems without claiming strict uniqueness: infinitely many nonlinear strictly increasing functions could, in principle, satisfy these moment constraints, but the anchor fixes the operationally relevant degrees of freedom for cross-system comparison.

This differs from post-hoc calibration, which rescales scores after training on a per-system basis and, therefore, cannot produce a universal numeric coordinate. Gauge fixing constrains the representation during learning, so the coordinate generalizes by construction.

\section{Benchmark System Equations and Parameters}
\label{app:systems}

For each system, we report the governing equations, fixed parameter values, the single parameter varied to traverse the predictability ladder, and the parameter range assigned to each regime. Regime boundaries were determined by the protocol described in Appendix~\ref{app:data}: the sign and relative magnitude of the leading Lyapunov exponent $\lambda_1$ cross-checked by visual inspection of attractor phase portraits. Initial conditions were sampled from multiple distributions per system (Gaussian perturbations about a base point, uniform on a sphere, and system-specific defaults); all distributions produced qualitatively consistent regime behavior.

The \emph{stable} range corresponds to L0, the \emph{transition} range to L1 (weakly chaotic), and the \emph{chaotic} range to L2 (strongly chaotic).

Unless noted otherwise, initial conditions were sampled from three distributions: Gaussian perturbations about a base point, uniform on a sphere, and system-specific defaults. All distributions produced qualitatively consistent regime behavior. Two exceptions are noted explicitly: Hastings--Powell, which requires ecologically structured non-negative initial conditions, and Chua, which additionally uses double-scroll sampling centered on the two attractor scrolls.

\subsection*{Source Systems (Training Set)}

\subsubsection*{Chen~\citep{chen1999yet}}

\begin{align}
\dot{x} &= a(y - x) \\
\dot{y} &= (c - a)x - xz + cy \\
\dot{z} &= xy - bz
\end{align}

Fixed parameters: $a = 35$, $b = 3$.
Varied parameter: $c$.

\begin{center}
\begin{tabular}{lc}
\toprule
Regime & $c$ range \\
\midrule
L0 (stable)     & $[15.0,\ 23.0)$ \\
L1 (transition) & $[23.0,\ 28.0)$ \\
L2 (chaotic)    & $[28.0,\ 33.0)$ \\
\bottomrule
\end{tabular}
\end{center}

\subsubsection*{Chua~\citep{chua1986double}}

\begin{align}
\dot{x} &= \alpha\bigl(y - x - f(x)\bigr) \\
\dot{y} &= x - y + z \\
\dot{z} &= -\beta y
\end{align}

where the piecewise-linear Chua diode is
\[
f(x) = m_1 x + \tfrac{1}{2}(m_0 - m_1)\bigl(|x+1| - |x-1|\bigr).
\]

Fixed parameters: $\beta = 28.0$, $m_0 = -1.143$, $m_1 = -0.714$.
Varied parameter: $\alpha$.

\begin{center}
\begin{tabular}{lc}
\toprule
Regime & $\alpha$ range \\
\midrule
L0 (stable)     & $[6.0,\ 10.0)$ \\
L1 (transition) & $[10.0,\ 13.0)$ \\
L2 (chaotic)    & $[13.0,\ 16.0)$ \\
\bottomrule
\end{tabular}
\end{center}

Chua additionally uses double-scroll sampling ($\sigma = 0.3$ about $(\pm 1.5,\ 0,\ 0)$) and Gaussian perturbations ($\sigma = 0.01$) about the base point $(0.7,\ 0,\ 0)$.

\subsubsection*{Duffing~\citep{duffing1918erzwungene}}

The Duffing oscillator is non-autonomous; the cosine forcing term renders the effective phase space three-dimensional.

\begin{align}
\dot{x} &= y \\
\dot{y} &= -\delta y - \alpha x - \beta x^3 + \gamma\cos(\omega t)
\end{align}

Fixed parameters: $\delta = 0.1$, $\alpha = -1.0$, $\beta = 1.0$,
$\omega = 1.0$.
Varied parameter: $\gamma$ (forcing amplitude).

\begin{center}
\begin{tabular}{lc}
\toprule
Regime & $\gamma$ range \\
\midrule
L0 (stable)     & $[0.05,\ 0.15)$ \\
L1 (transition) & $[0.25,\ 0.35)$ \\
L2 (chaotic)    & $[0.35,\ 0.60)$ \\
\bottomrule
\end{tabular}
\end{center}

\subsubsection*{Finance~\citep{cai2007new}}

\begin{align}
\dot{x} &= \left(\tfrac{1}{b} - a\right)x + z + xy \\
\dot{y} &= -by - x^2 \\
\dot{z} &= -x - cz
\end{align}

Fixed parameters: $b = 0.2$, $c = 1.1$.
Varied parameter: $a$.

\begin{center}
\begin{tabular}{lc}
\toprule
Regime & $a$ range \\
\midrule
L0 (stable)     & $(4.0,\ 8.0]$ \\
L1 (transition) & $[3.0,\ 4.0)$ \\
L2 (chaotic)    & $[0.001,\ 3.0)$ \\
\bottomrule
\end{tabular}
\end{center}

\subsubsection*{Genesio--Tesi~\citep{genesio1992harmonic}}

\begin{align}
\dot{x} &= y \\
\dot{y} &= z \\
\dot{z} &= -cx - by - az + x^2
\end{align}

Fixed parameters: $b = 1.1$, $c = 1.0$.
Varied parameter: $a$.

\begin{center}
\begin{tabular}{lc}
\toprule
Regime & $a$ range \\
\midrule
L0 (stable)     & $[0.65,\ 1.20)$ \\
L1 (transition) & $[0.50,\ 0.65)$ \\
L2 (chaotic)    & $[0.20,\ 0.50)$ \\
\bottomrule
\end{tabular}
\end{center}

\subsubsection*{Halvorsen~\citep{sprott2010elegant}}

\begin{align}
\dot{x} &= -ax - by - bz - y^2 \\
\dot{y} &= -ay - bz - bx - z^2 \\
\dot{z} &= -az - bx - by - x^2
\end{align}

Fixed parameter: $b = 4.0$.
Varied parameter: $a$.

\begin{center}
\begin{tabular}{lc}
\toprule
Regime & $a$ range \\
\midrule
L0 (stable)     & $[1.86,\ 4.0)$ \\
L1 (transition) & $[1.80,\ 1.86)$ \\
L2 (chaotic)    & $[1.20,\ 1.80)$ \\
\bottomrule
\end{tabular}
\end{center}

\subsubsection*{Hastings--Powell~\citep{hastings1991chaos}}

\begin{align}
\dot{x} &= x(1-x) - \frac{a_1 xy}{1 + b_1 x} \\
\dot{y} &= \frac{a_1 xy}{1 + b_1 x} - \frac{a_2 yz}{1 + b_2 y} - d_1 y \\
\dot{z} &= \frac{a_2 yz}{1 + b_2 y} - d_2 z
\end{align}

Fixed parameters: $a_2 = 0.1$, $b_1 = 3.0$, $b_2 = 2.0$, $d_1 = 0.4$, $d_2 = 0.01$. Varied parameter: $a_1$ (prey--predator interaction rate).

\begin{center}
\begin{tabular}{lc}
\toprule
Regime & $a_1$ range \\
\midrule
L0 (stable)     & $[2.5,\ 3.5)$ \\
L1 (transition) & $[3.5,\ 4.8)$ \\
L2 (chaotic)    & $[4.8,\ 5.2)$ \\
\bottomrule
\end{tabular}
\end{center}

Initial conditions were absolute-valued to ensure non-negative populations, with ecological structured sampling (prey $\in [0.5,1.0]$, predator $\in [0.1,0.4]$, superpredator $\in [0.2,0.8]$) used in place of sphere sampling. Integration used an adaptive timescale with an $8\times$ ecological multiplier applied to the estimated dominant period $\tau$, reflecting the longer characteristic timescales of trophic oscillations.

\subsubsection*{Lorenz-63~\citep{lorenz1963deterministic}}

\begin{align}
\dot{x} &= \sigma(y - x) \\
\dot{y} &= x(\rho - z) - y \\
\dot{z} &= xy - \beta z
\end{align}

Fixed parameters: $\sigma = 10.0$, $\beta = 8/3$. Varied parameter: $\rho$.

\begin{center}
\begin{tabular}{lc}
\toprule
Regime & $\rho$ range \\
\midrule
L0 (stable)     & $[10.0,\ 20.0)$ \\
L1 (transition) & $[20.0,\ 28.0)$ \\
L2 (chaotic)    & $[28.0,\ 40.0)$ \\
\bottomrule
\end{tabular}
\end{center}

\subsubsection*{Lorenz-84~\citep{lorenz1984irregularity}}

\begin{align}
\dot{x} &= -ax - y^2 - z^2 + aF \\
\dot{y} &= -y + xy - bxz + G \\
\dot{z} &= -z + bxy + xz
\end{align}

Fixed parameters: $a = 0.25$, $b = 4.0$, $G = 1.0$. Varied parameter: $F$ (forcing).

\begin{center}
\begin{tabular}{lc}
\toprule
Regime & $F$ range \\
\midrule
L0 (stable)     & $[1.5,\ 4.0)$ \\
L1 (transition) & $[4.0,\ 6.0)$ \\
L2 (chaotic)    & $[6.0,\ 10.0)$ \\
\bottomrule
\end{tabular}
\end{center}

\subsubsection*{R\"{o}ssler~\citep{rossler1976equation}}

\begin{align}
\dot{x} &= -y - z \\
\dot{y} &= x + ay \\
\dot{z} &= b + z(x - c)
\end{align}

Fixed parameters: $a = 0.2$, $b = 0.2$. Varied parameter: $c$.

\begin{center}
\begin{tabular}{lc}
\toprule
Regime & $c$ range \\
\midrule
L0 (stable)     & $[2.0,\ 2.83)$ \\
L1 (transition) & $[2.83,\ 4.2)$ \\
L2 (chaotic)    & $[4.2,\ 6.0)$ \\
\bottomrule
\end{tabular}
\end{center}

\subsubsection*{Rucklidge~\citep{Rucklidge_1992}}

\begin{align}
\dot{x} &= -ax + by - yz \\
\dot{y} &= x \\
\dot{z} &= -z + y^2
\end{align}

Fixed parameter: $a = 2.0$. Varied parameter: $b$.

\begin{center}
\begin{tabular}{lc}
\toprule
Regime & $b$ range \\
\midrule
L0 (stable)     & $[1.0,\ 3.0)$ \\
L1 (transition) & $[3.0,\ 5.5)$ \\
L2 (chaotic)    & $[5.5,\ 8.0)$ \\
\bottomrule
\end{tabular}
\end{center}

\subsubsection*{Shimizu--Morioka~\citep{shimizu1980bifurcation}}

\begin{align}
\dot{x} &= y \\
\dot{y} &= x - ay - xz \\
\dot{z} &= -bz + x^2
\end{align}

Fixed parameter: $b = 0.5$.
Varied parameter: $a$.

\begin{center}
\begin{tabular}{lc}
\toprule
Regime & $a$ range \\
\midrule
L0 (stable)     & $[1.30,\ 1.60)$ \\
L1 (transition) & $[1.20,\ 1.30)$ \\
L2 (chaotic)    & $[0.10,\ 1.20)$ \\
\bottomrule
\end{tabular}
\end{center}

\subsection*{Target Systems (Held-Out Set)}

\subsubsection*{Forced Pendulum~\citep{d1982chaotic}}

Non-autonomous; cosine forcing renders the effective phase space three-dimensional.

\begin{align}
\dot{\theta} &= v \\
\dot{v}      &= -\gamma v - \sin\theta + A\cos(\omega t)
\end{align}

Fixed parameters: $\gamma = 0.2$, $\omega = 2/3$.
Varied parameter: $A$ (forcing amplitude).

\begin{center}
\begin{tabular}{lc}
\toprule
Regime & $A$ range \\
\midrule
L0 (stable)     & $[0.30,\ 0.60)$ \\
L1 (transition) & $[0.60,\ 0.85)$ \\
L2 (chaotic)    & $[0.90,\ 1.40)$ \\
\bottomrule
\end{tabular}
\end{center}

\subsubsection*{Newton--Leipnik~\citep{Leipnik_1981}}

\begin{align}
\dot{x} &= -ax + y + 10yz \\
\dot{y} &= -x - 0.4y + 5xz \\
\dot{z} &= bz - 5xy
\end{align}

Fixed parameter: $a = 0.4$. Varied parameter: $b$ (z-axis damping).

\begin{center}
\begin{tabular}{lc}
\toprule
Regime & $b$ range \\
\midrule
L0 (stable)     & $[0.030,\ 0.100)$ \\
L1 (transition) & $[0.100,\ 0.150)$ \\
L2 (chaotic)    & $[0.150,\ 0.200)$ \\
\bottomrule
\end{tabular}
\end{center}

\subsubsection*{Rabinovich--Fabrikant~\citep{rabinovich1979stochastic}}

\begin{align}
\dot{x} &= y(z - 1 + x^2) + \gamma x \\
\dot{y} &= x(3z + 1 - x^2) + \gamma y \\
\dot{z} &= -2z(\alpha + xy)
\end{align}

Fixed parameter: $\alpha = 0.1$. Varied parameter: $\gamma$.

\begin{center}
\begin{tabular}{lc}
\toprule
Regime & $\gamma$ range \\
\midrule
L0 (stable)     & $[0.05,\ 0.12)$ \\
L1 (transition) & $[0.12,\ 0.26)$ \\
L2 (chaotic)    & $[0.26,\ 0.29)$ \\
\bottomrule
\end{tabular}
\end{center}

Trajectories exceeding $\|s\|_\infty = 25$ were truncated to the bounded portion before divergence; trajectories with fewer than
300 retained points were discarded.

\subsubsection*{Thomas~\citep{thomas1999deterministic}}

\begin{align}
\dot{x} &= \sin y - bx \\
\dot{y} &= \sin z - by \\
\dot{z} &= \sin x - bz
\end{align}

No fixed parameters. Varied parameter: $b$ (dissipation).

\begin{center}
\begin{tabular}{lc}
\toprule
Regime & $b$ range \\
\midrule
L0 (stable)     & $[0.25,\ 0.35)$ \\
L1 (transition) & $[0.15,\ 0.25)$ \\
L2 (chaotic)    & $[0.05,\ 0.15)$ \\
\bottomrule
\end{tabular}
\end{center}

\subsubsection*{Tigan~\citep{Tigan_2008}}

\begin{align}
\dot{x} &= a(y - x) \\
\dot{y} &= (c - a)x - axz \\
\dot{z} &= -bz + xy
\end{align}

Fixed parameters: $a = 2.1$, $b = 0.6$. Varied parameter: $c$.

\begin{center}
\begin{tabular}{lc}
\toprule
Regime & $c$ range \\
\midrule
L0 (stable)     & $[5.0,\ 12.0)$ \\
L1 (transition) & $[12.0,\ 22.0)$ \\
L2 (chaotic)    & $[22.0,\ 32.0)$ \\
\bottomrule
\end{tabular}
\end{center}

\section{Neural Baseline Comparison}
\label{app:neural}

We compare GON with three neural baselines trained on the same 12 source systems using identical data, preprocessing, encoders, and optimization protocols. The only change is the training objective.

\textbf{Regression} minimizes mean squared error to integer labels $\{0,1,2,3,4\}$.
\textbf{Classification} minimizes cross-entropy over five classes.
\textbf{CORAL}~\citep{cao2020rank} uses cumulative link constraints for ordinal supervision.

Each method is converted to a scalar score for evaluation: the raw output for regression, the expected label $\sum_k k\,p_k$ for classification, and the cumulative-logit score for CORAL.

\paragraph{In-distribution.}
Table~\ref{tab:neural_indist} reports test performance on the 12 source systems. Regression, classification, CORAL, and GON all achieve near-perfect adjacent-pair discrimination in-distribution. To make cross-method comparisons meaningful despite different score scales, we report the scale-normalized monotonicity slope $\beta_{\text{norm}}$ rather than raw $\beta$. Under this normalization, regression, classification, CORAL, and GON all show nearly perfect global ordering on the source distribution. This pattern indicates that source-domain performance alone is insufficient to reveal whether a learned score will transfer with a stable numerical meaning.

\begin{table}[ht]
\centering
\caption{Neural baselines --- in-distribution performance on the 12 source systems (test split).}
\label{tab:neural_indist}
\small
\begin{tabular}{lrrrrrr}
\toprule
Method & $\beta_{\text{norm}}$ & $\auroc{01}$ & $\auroc{12}$
       & $\auroc{23}$ & $\auroc{34}$ & $\bar{\sigma}$ \\
\midrule
Regression      & 0.995 & 0.998 & 1.000 & 0.996 & 1.000 & 0.096 \\
Classification  & 0.993 & 0.998 & 0.998 & 0.997 & 1.000 & 0.087 \\
CORAL           & 0.994 & 0.999 & 0.999 & 0.996 & 1.000 & 0.091 \\
\midrule
\textbf{GON}    & \textbf{0.999} & 0.998 & 1.000 & 0.998 & 1.000 & 0.191 \\
\bottomrule
\end{tabular}
\end{table}

\paragraph{Zero-shot transfer.}
Table~\ref{tab:neural_Transfer} reports zero-shot performance on the 5 held-out target systems. All supervised baselines cluster at similar values of $\beta_{\text{norm}} \approx 0.46$--$0.48$ despite their different objectives, suggesting a shared limitation: none learns a score with a stable cross-system coordinate. 

GON achieves the highest scale-normalized slope, $\beta_{\text{norm}}=0.559$, while also remaining competitive on the strongest transferable boundary $L_2 \rightarrow L_3$. This is the relevant comparison for cross-system ordinal structure, since raw $\beta$ is not directly comparable across methods with different score scales.

\begin{table}[ht]
\centering
\caption{Neural baselines --- zero-shot transfer on the 5 held-out systems (macro-averaged over per-system values).}
\label{tab:neural_Transfer}
\small
\begin{tabular}{lrrrrrr}
\toprule
Method & $\beta_{\text{norm}}$
       & $\auroc{01}$ & $\auroc{12}$ & $\auroc{23}$ & $\auroc{34}$ \\
\midrule
Regression     & 0.479 & 0.501 & 0.636 & 0.643 & 0.980 \\
Classification & 0.463 & 0.640 & 0.541 & 0.617 & 0.984 \\
CORAL          & 0.461 & 0.648 & 0.606 & 0.724 & 0.992 \\
\midrule
\textbf{GON}   & \textbf{0.559} & 0.494 & 0.583 & 0.715 & 0.973 \\
\bottomrule
\end{tabular}
\end{table}

\section{Per-System Adaptation Results}
\label{app:adaptation}
Tables~\ref{tab:app_nl_mono}--\ref{tab:app_tigan_auroc} report monotonicity 
slope $\beta$ and all four adjacent-pair AUROC values for each of the five 
held-out systems, across $k \in \{5,10,\ldots,100\}$ labeled windows. All 
entries are mean\,$\pm$\, std over 5 random seeds. Zero-shot (ZS) performance 
is listed in the caption of each system block for reference. Columns correspond 
to the three adaptation strategies: \textbf{scratch} (random initialization), 
\textbf{pre\_head} (frozen encoder, readout fine-tuned), and \textbf{pre\_all} 
(full model fine-tuned from the pretrained checkpoint).

\paragraph{Fine-tuning protocol.}
All three adaptation strategies use AdamW with learning rate $2\times10^{-3}$, 
weight decay $10^{-4}$, cosine annealing over 30 epochs, and gradient clipping 
at 5.0; identical to pretraining. Data augmentation and EMA are disabled during 
fine-tuning. For {\bf pre\_head}, all encoder parameters are frozen including 
GroupNorm weight/bias; only the readout, normalization layer, and scale/bias 
embeddings are updated. The trajectory-level train/val/test split is fixed 
across all runs (seed 42). For each $(k, \text{seed})$ pair, $k$ windows are 
drawn without replacement from the fixed training pool, with at least one 
window per ladder level before filling the remaining budget uniformly at random.

\newcommand{\apptab}[1]{\small #1}

\subsection*{Newton--Leipnik}
\textbf{Zero-shot:}
$\beta=0.737$,
$\auroc{01}=0.539$,
$\auroc{12}=0.514$,
$\auroc{23}=0.856$,
$\auroc{34}=1.000$.

\begin{table}[ht]
\centering\small
\caption{Newton--Leipnik: monotonicity slope $\beta$ (mean\,$\pm$\,std, 5 seeds).}
\label{tab:app_nl_mono}
\begin{tabular}{r lll}
\toprule
$k$ & scratch & pre\_head & pre\_all \\
\midrule
5   & 0.730$\pm$0.270 & 0.852$\pm$0.067 & \textbf{1.328$\pm$0.263} \\
10  & 0.650$\pm$0.300 & 0.892$\pm$0.056 & \textbf{1.443$\pm$0.223} \\
15  & 0.954$\pm$0.180 & 0.938$\pm$0.065 & \textbf{1.541$\pm$0.198} \\
20  & 0.942$\pm$0.157 & 0.950$\pm$0.086 & \textbf{1.496$\pm$0.167} \\
25  & 0.785$\pm$0.303 & 0.917$\pm$0.038 & \textbf{1.415$\pm$0.285} \\
30  & 0.765$\pm$0.400 & 0.959$\pm$0.026 & \textbf{1.615$\pm$0.163} \\
35  & 0.830$\pm$0.268 & 0.959$\pm$0.055 & \textbf{1.592$\pm$0.142} \\
40  & 1.157$\pm$0.225 & 0.935$\pm$0.083 & \textbf{1.588$\pm$0.146} \\
45  & 0.903$\pm$0.294 & 0.910$\pm$0.131 & \textbf{1.649$\pm$0.219} \\
50  & 1.032$\pm$0.253 & 0.996$\pm$0.020 & \textbf{1.636$\pm$0.129} \\
55  & 0.611$\pm$0.232 & 0.908$\pm$0.084 & \textbf{1.782$\pm$0.084} \\
60  & 0.956$\pm$0.327 & 0.971$\pm$0.066 & \textbf{1.568$\pm$0.221} \\
65  & 0.904$\pm$0.441 & 0.940$\pm$0.073 & \textbf{1.666$\pm$0.180} \\
70  & 1.109$\pm$0.296 & 1.018$\pm$0.023 & \textbf{1.597$\pm$0.141} \\
75  & 1.127$\pm$0.376 & 0.977$\pm$0.021 & \textbf{1.658$\pm$0.077} \\
80  & 0.919$\pm$0.299 & 0.932$\pm$0.072 & \textbf{1.664$\pm$0.148} \\
85  & 0.965$\pm$0.213 & 0.943$\pm$0.047 & \textbf{1.675$\pm$0.134} \\
90  & 1.011$\pm$0.248 & 0.965$\pm$0.035 & \textbf{1.703$\pm$0.135} \\
95  & 0.857$\pm$0.357 & 0.974$\pm$0.063 & \textbf{1.654$\pm$0.087} \\
100 & 1.133$\pm$0.148 & 0.974$\pm$0.029 & \textbf{1.620$\pm$0.104} \\
\bottomrule
\end{tabular}
\end{table}

\begin{table}[ht]
\centering\small
\caption{Newton--Leipnik: $\auroc{01}$, $\auroc{12}$, $\auroc{23}$, $\auroc{34}$ (mean\,$\pm$\,std, 5 seeds).}
\label{tab:app_nl_auroc}
\begin{tabular}{r lll lll lll lll}
\toprule
& \multicolumn{3}{c}{$\auroc{01}$}
& \multicolumn{3}{c}{$\auroc{12}$}
& \multicolumn{3}{c}{$\auroc{23}$}
& \multicolumn{3}{c}{$\auroc{34}$} \\
\cmidrule(lr){2-4}\cmidrule(lr){5-7}\cmidrule(lr){8-10}\cmidrule(lr){11-13}
$k$ & scr & ph & pa & scr & ph & pa & scr & ph & pa & scr & ph & pa \\
\midrule
5   & .620 & .581 & .619 & .609 & .502 & .480 & .486 & .903 & \textbf{.919} & .911 & \textbf{1.00} & .997 \\
10  & .668 & .589 & .636 & .596 & .506 & .536 & .467 & .901 & \textbf{.927} & .887 & \textbf{1.00} & .985 \\
15  & .656 & .594 & .643 & .630 & .506 & .526 & .465 & .907 & \textbf{.900} & .960 & \textbf{1.00} & .997 \\
20  & .749 & .591 & .639 & .641 & .508 & .548 & .435 & .920 & \textbf{.941} & .995 & \textbf{1.00} & .983 \\
25  & .714 & .603 & .604 & .630 & .506 & .526 & .451 & .919 & \textbf{.907} & .989 & \textbf{1.00} & .973 \\
30  & .732 & .599 & .663 & .630 & .510 & .547 & .461 & .930 & \textbf{.921} & .998 & \textbf{1.00} & .974 \\
35  & .748 & .601 & .641 & .647 & .507 & .525 & .458 & .929 & \textbf{.961} & .998 & \textbf{1.00} & .994 \\
40  & .739 & .607 & .677 & .646 & .506 & .520 & .443 & .931 & \textbf{.893} & .998 & \textbf{1.00} & .978 \\
45  & .739 & .594 & .677 & .654 & .506 & .563 & .432 & .922 & \textbf{.921} & .995 & \textbf{1.00} & .975 \\
50  & .725 & .604 & .669 & .638 & .510 & .550 & .455 & .933 & \textbf{.923} & .990 & \textbf{1.00} & .980 \\
55  & .716 & .608 & .678 & .643 & .503 & .536 & .452 & .922 & \textbf{.888} & .974 & \textbf{1.00} & .957 \\
60  & .756 & .603 & .681 & .663 & .512 & .548 & .439 & .923 & \textbf{.901} & .998 & \textbf{1.00} & .977 \\
65  & .737 & .601 & .672 & .636 & .505 & .527 & .463 & .921 & \textbf{.898} & .997 & \textbf{1.00} & .980 \\
70  & .747 & .600 & .677 & .647 & .509 & .571 & .457 & .931 & \textbf{.938} & .999 & \textbf{1.00} & .978 \\
75  & .756 & .598 & .670 & .663 & .512 & .584 & .445 & .925 & \textbf{.918} & .998 & \textbf{1.00} & .976 \\
80  & .734 & .603 & .687 & .648 & .505 & .566 & .449 & .920 & \textbf{.925} & .993 & \textbf{1.00} & .976 \\
85  & .749 & .594 & .688 & .652 & .510 & .616 & .450 & .918 & \textbf{.892} & .998 & \textbf{1.00} & .954 \\
90  & .726 & .606 & .701 & .633 & .512 & .558 & .463 & .927 & \textbf{.904} & .997 & \textbf{1.00} & .974 \\
95  & .735 & .606 & .675 & .653 & .509 & .598 & .447 & .931 & \textbf{.897} & .998 & \textbf{1.00} & .986 \\
100 & .738 & .605 & .690 & .632 & .511 & .538 & .459 & .929 & \textbf{.932} & .997 & \textbf{1.00} & .984 \\
\bottomrule
\end{tabular}
\end{table}

\subsection*{Forced Pendulum}
\textbf{Zero-shot:}
$\beta=0.489$,
$\auroc{01}=0.295$,
$\auroc{12}=0.390$,
$\auroc{23}=0.604$,
$\auroc{34}=1.000$.

\begin{table}[ht]
\centering\small
\caption{Forced Pendulum: monotonicity slope $\beta$ (mean\,$\pm$\,std, 5 seeds).}
\label{tab:app_pend_mono}
\begin{tabular}{r lll}
\toprule
$k$ & scratch & pre\_head & pre\_all \\
\midrule
5   & 0.110$\pm$0.072 & 0.609$\pm$0.035 & \textbf{0.701$\pm$0.067} \\
10  & 0.213$\pm$0.190 & 0.606$\pm$0.023 & \textbf{0.780$\pm$0.064} \\
15  & 0.338$\pm$0.368 & 0.599$\pm$0.060 & \textbf{0.775$\pm$0.169} \\
20  & 0.374$\pm$0.451 & 0.596$\pm$0.040 & \textbf{0.733$\pm$0.157} \\
25  & 0.606$\pm$0.505 & 0.633$\pm$0.028 & \textbf{0.834$\pm$0.117} \\
30  & 0.747$\pm$0.392 & 0.634$\pm$0.020 & \textbf{0.722$\pm$0.191} \\
35  & 0.599$\pm$0.422 & 0.616$\pm$0.044 & \textbf{0.981$\pm$0.116} \\
40  & 0.522$\pm$0.188 & 0.624$\pm$0.056 & \textbf{0.973$\pm$0.189} \\
45  & 0.522$\pm$0.251 & 0.646$\pm$0.039 & \textbf{0.880$\pm$0.113} \\
50  & 0.378$\pm$0.350 & 0.649$\pm$0.034 & \textbf{1.084$\pm$0.150} \\
55  & 0.730$\pm$0.389 & 0.654$\pm$0.041 & \textbf{1.039$\pm$0.237} \\
60  & 0.546$\pm$0.440 & 0.645$\pm$0.016 & \textbf{0.876$\pm$0.272} \\
65  & 0.662$\pm$0.480 & 0.646$\pm$0.034 & \textbf{1.107$\pm$0.185} \\
70  & 0.553$\pm$0.552 & 0.625$\pm$0.033 & \textbf{1.050$\pm$0.163} \\
75  & 0.686$\pm$0.615 & 0.656$\pm$0.030 & \textbf{0.937$\pm$0.209} \\
80  & 0.746$\pm$0.345 & 0.658$\pm$0.014 & \textbf{0.880$\pm$0.174} \\
85  & 0.740$\pm$0.458 & 0.664$\pm$0.017 & \textbf{0.960$\pm$0.224} \\
90  & 0.562$\pm$0.554 & 0.648$\pm$0.028 & \textbf{1.035$\pm$0.203} \\
95  & 0.403$\pm$0.287 & 0.633$\pm$0.032 & \textbf{0.921$\pm$0.102} \\
100 & 0.589$\pm$0.496 & 0.636$\pm$0.026 & \textbf{0.835$\pm$0.135} \\
\bottomrule
\end{tabular}
\end{table}

\begin{table}[ht]
\centering\small
\caption{Forced Pendulum: $\auroc{01}$, $\auroc{12}$, $\auroc{23}$, $\auroc{34}$ (mean\,$\pm$\,std, 5 seeds). Note: $\auroc{23}$ does not improve beyond zero-shot (0.604) for any method or any $k$, indicating that the $L_2\!\to\!L_3$ boundary is not resolved by window-level adaptation alone.}
\label{tab:app_pend_auroc}
\begin{tabular}{r lll lll lll lll}
\toprule
& \multicolumn{3}{c}{$\auroc{01}$}
& \multicolumn{3}{c}{$\auroc{12}$}
& \multicolumn{3}{c}{$\auroc{23}$}
& \multicolumn{3}{c}{$\auroc{34}$} \\
\cmidrule(lr){2-4}\cmidrule(lr){5-7}\cmidrule(lr){8-10}\cmidrule(lr){11-13}
$k$ & scr & ph & pa & scr & ph & pa & scr & ph & pa & scr & ph & pa \\
\midrule
5   & .390 & .337 & \textbf{.460} & .485 & .448 & .450 & .534 & .593 & \textbf{.604} & .699 & \textbf{1.00} & 1.00 \\
10  & .434 & .447 & \textbf{.638} & .492 & .431 & .446 & .507 & .566 & \textbf{.551} & .741 & \textbf{1.00} & 1.00 \\
15  & .424 & .337 & \textbf{.566} & .484 & .471 & .480 & .526 & \textbf{.610} & .600 & .801 & \textbf{1.00} & .996 \\
20  & .593 & .353 & \textbf{.670} & .456 & .437 & .456 & .512 & \textbf{.593} & .543 & .703 & \textbf{1.00} & .998 \\
25  & .565 & .464 & \textbf{.641} & .526 & .501 & .513 & .478 & .550 & \textbf{.551} & .783 & \textbf{1.00} & 1.00 \\
30  & .654 & .544 & \textbf{.644} & .509 & .501 & .487 & .465 & .518 & \textbf{.526} & .703 & \textbf{1.00} & .992 \\
35  & .515 & .405 & \textbf{.624} & .514 & .531 & .501 & .516 & .577 & \textbf{.625} & .751 & \textbf{1.00} & .998 \\
40  & .639 & .472 & \textbf{.728} & .497 & .456 & .486 & .474 & .542 & \textbf{.569} & .731 & \textbf{1.00} & .995 \\
45  & .622 & .479 & \textbf{.627} & .513 & .471 & .522 & .493 & .552 & \textbf{.555} & .720 & \textbf{1.00} & 1.00 \\
50  & .452 & .351 & \textbf{.644} & .479 & .457 & .517 & .509 & .585 & \textbf{.626} & .810 & \textbf{1.00} & .999 \\
55  & .605 & .437 & \textbf{.630} & .530 & .466 & .498 & .483 & .566 & \textbf{.605} & .740 & \textbf{1.00} & .991 \\
60  & .543 & .476 & \textbf{.638} & .539 & .482 & .496 & .498 & .552 & \textbf{.584} & .737 & \textbf{1.00} & 1.00 \\
65  & .542 & .485 & \textbf{.691} & .558 & .455 & .534 & .493 & .552 & \textbf{.571} & .675 & \textbf{1.00} & 1.00 \\
70  & .623 & .505 & \textbf{.661} & .468 & .472 & .495 & .500 & .522 & \textbf{.609} & .714 & \textbf{1.00} & .993 \\
75  & .658 & .513 & \textbf{.706} & .528 & .511 & .532 & .492 & .539 & \textbf{.580} & .713 & \textbf{1.00} & .997 \\
80  & .628 & .398 & \textbf{.556} & .535 & .514 & .550 & .517 & .579 & \textbf{.618} & .758 & \textbf{1.00} & .995 \\
85  & .564 & .407 & \textbf{.647} & .554 & .508 & .523 & .484 & .577 & \textbf{.613} & .777 & \textbf{1.00} & .998 \\
90  & .618 & .445 & \textbf{.747} & .517 & .474 & .526 & .506 & .559 & \textbf{.581} & .695 & \textbf{1.00} & .992 \\
95  & .639 & .506 & \textbf{.645} & .531 & .488 & .487 & .481 & .542 & \textbf{.581} & .744 & \textbf{1.00} & .994 \\
100 & .623 & .412 & \textbf{.637} & .546 & .487 & .493 & .473 & .582 & \textbf{.595} & .806 & \textbf{1.00} & 1.00 \\
\bottomrule
\end{tabular}
\end{table}

\subsection*{Rabinovich--Fabrikant}
\textbf{Zero-shot:}
$\beta=1.270$,
$\auroc{01}=0.483$,
$\auroc{12}=0.576$,
$\auroc{23}=0.725$,
$\auroc{34}=0.872$.

\begin{table}[ht]
\centering\small
\caption{Rabinovich--Fabrikant: monotonicity slope $\beta$ (mean\,$\pm$\,std, 5 seeds). \textbf{scratch} fails to recover ordinal structure across the entire $k$ range, never exceeding $\beta=0.84$ even at $k=100$.}
\label{tab:app_rf_mono}
\begin{tabular}{r lll}
\toprule
$k$ & scratch & pre\_head & pre\_all \\
\midrule
5   & 0.009$\pm$0.050 & 1.153$\pm$0.058 & \textbf{1.256$\pm$0.143} \\
10  & 0.218$\pm$0.278 & 1.203$\pm$0.075 & \textbf{1.400$\pm$0.250} \\
15  & 0.254$\pm$0.315 & 1.179$\pm$0.016 & \textbf{1.345$\pm$0.211} \\
20  & 0.244$\pm$0.209 & 1.281$\pm$0.075 & \textbf{1.503$\pm$0.110} \\
25  & 0.280$\pm$0.255 & 1.263$\pm$0.130 & \textbf{1.402$\pm$0.255} \\
30  & 0.277$\pm$0.206 & 1.226$\pm$0.106 & \textbf{1.460$\pm$0.239} \\
35  & 0.128$\pm$0.105 & 1.274$\pm$0.126 & \textbf{1.550$\pm$0.175} \\
40  & 0.180$\pm$0.237 & 1.247$\pm$0.080 & \textbf{1.506$\pm$0.192} \\
45  & 0.468$\pm$0.478 & 1.226$\pm$0.113 & \textbf{1.480$\pm$0.207} \\
50  & 0.536$\pm$0.409 & 1.270$\pm$0.076 & \textbf{1.519$\pm$0.184} \\
55  & 0.430$\pm$0.399 & 1.240$\pm$0.094 & \textbf{1.682$\pm$0.128} \\
60  & 0.155$\pm$0.285 & 1.394$\pm$0.065 & \textbf{1.639$\pm$0.114} \\
65  & 0.410$\pm$0.373 & 1.379$\pm$0.061 & \textbf{1.609$\pm$0.197} \\
70  & 0.537$\pm$0.418 & 1.352$\pm$0.068 & \textbf{1.720$\pm$0.080} \\
75  & 0.425$\pm$0.488 & 1.295$\pm$0.073 & \textbf{1.519$\pm$0.200} \\
80  & 0.601$\pm$0.453 & 1.289$\pm$0.123 & \textbf{1.644$\pm$0.202} \\
85  & 0.421$\pm$0.280 & 1.329$\pm$0.044 & \textbf{1.621$\pm$0.192} \\
90  & 0.836$\pm$0.414 & 1.344$\pm$0.041 & \textbf{1.598$\pm$0.158} \\
95  & 0.379$\pm$0.250 & 1.251$\pm$0.057 & \textbf{1.445$\pm$0.187} \\
100 & 0.105$\pm$0.115 & 1.282$\pm$0.092 & \textbf{1.591$\pm$0.175} \\
\bottomrule
\end{tabular}
\end{table}

\begin{table}[ht]
\centering\small
\caption{Rabinovich--Fabrikant: $\auroc{01}$, $\auroc{12}$, $\auroc{23}$, $\auroc{34}$ (mean\,$\pm$\,std, 5 seeds).}
\label{tab:app_rf_auroc}
\begin{tabular}{r lll lll lll lll}
\toprule
& \multicolumn{3}{c}{$\auroc{01}$}
& \multicolumn{3}{c}{$\auroc{12}$}
& \multicolumn{3}{c}{$\auroc{23}$}
& \multicolumn{3}{c}{$\auroc{34}$} \\
\cmidrule(lr){2-4}\cmidrule(lr){5-7}\cmidrule(lr){8-10}\cmidrule(lr){11-13}
$k$ & scr & ph & pa & scr & ph & pa & scr & ph & pa & scr & ph & pa \\
\midrule
5   & .428 & .490 & \textbf{.519} & .554 & .576 & \textbf{.557} & .498 & .725 & \textbf{.766} & .455 & .883 & \textbf{.895} \\
10  & .484 & .492 & \textbf{.541} & .480 & .575 & \textbf{.568} & .523 & .726 & \textbf{.780} & .602 & .875 & \textbf{.900} \\
15  & .495 & .496 & \textbf{.519} & .509 & .578 & \textbf{.573} & .541 & .727 & \textbf{.757} & .524 & .879 & \textbf{.905} \\
20  & .552 & .494 & \textbf{.514} & .519 & .581 & \textbf{.579} & .525 & .734 & \textbf{.812} & .562 & .883 & \textbf{.919} \\
25  & .548 & .495 & \textbf{.523} & .484 & .578 & \textbf{.570} & .590 & .731 & \textbf{.769} & .517 & .870 & \textbf{.885} \\
30  & .535 & .483 & \textbf{.536} & .452 & .577 & \textbf{.562} & .589 & .723 & \textbf{.792} & .597 & .869 & \textbf{.906} \\
35  & .522 & .501 & \textbf{.520} & .495 & .579 & \textbf{.571} & .556 & .740 & \textbf{.819} & .529 & .878 & \textbf{.905} \\
40  & .485 & .495 & \textbf{.538} & .546 & .580 & \textbf{.580} & .521 & .739 & \textbf{.795} & .529 & .880 & \textbf{.914} \\
45  & .600 & .494 & \textbf{.501} & .488 & .581 & \textbf{.590} & .633 & .727 & \textbf{.791} & .535 & .898 & \textbf{.919} \\
50  & .543 & .498 & \textbf{.529} & .465 & .581 & \textbf{.581} & .590 & .738 & \textbf{.798} & .614 & .887 & \textbf{.903} \\
55  & .548 & .510 & \textbf{.558} & .500 & .581 & \textbf{.584} & .588 & .736 & \textbf{.825} & .561 & .886 & \textbf{.916} \\
60  & .491 & .500 & \textbf{.524} & .477 & .582 & \textbf{.587} & .538 & .739 & \textbf{.807} & .572 & .885 & \textbf{.909} \\
65  & .558 & .505 & \textbf{.583} & .493 & .579 & \textbf{.581} & .604 & .736 & \textbf{.816} & .546 & .884 & \textbf{.907} \\
70  & .543 & .506 & \textbf{.577} & .515 & .581 & \textbf{.584} & .624 & .740 & \textbf{.828} & .518 & .886 & \textbf{.916} \\
75  & .540 & .493 & \textbf{.523} & .522 & .582 & \textbf{.594} & .557 & .736 & \textbf{.778} & .547 & .880 & \textbf{.921} \\
80  & .570 & .501 & \textbf{.570} & .481 & .580 & \textbf{.567} & .613 & .735 & \textbf{.822} & .597 & .883 & \textbf{.908} \\
85  & .583 & .500 & \textbf{.560} & .523 & .584 & \textbf{.586} & .574 & .737 & \textbf{.799} & .574 & .887 & \textbf{.911} \\
90  & .579 & .498 & \textbf{.534} & .481 & .582 & \textbf{.596} & .638 & .739 & \textbf{.799} & .627 & .885 & \textbf{.924} \\
95  & .546 & .503 & \textbf{.542} & .480 & .578 & \textbf{.576} & .590 & .734 & \textbf{.781} & .569 & .880 & \textbf{.902} \\
100 & .511 & .499 & \textbf{.555} & .502 & .578 & \textbf{.586} & .511 & .730 & \textbf{.789} & .574 & .881 & \textbf{.908} \\
\bottomrule
\end{tabular}
\end{table}

\subsection*{Thomas}
\textbf{Zero-shot:}
$\beta=0.958$,
$\auroc{01}=0.508$,
$\auroc{12}=0.452$,
$\auroc{23}=0.391$,
$\auroc{34}=0.995$.

\begin{table}[ht]
\centering\small
\caption{Thomas: monotonicity slope $\beta$ (mean\,$\pm$\,std, 5 seeds).
\textbf{pre\_all} beats zero-shot at most $k$ values but with high variance;
\textbf{scratch} becomes competitive on $\beta$ above $k=40$ while remaining
below \textbf{pre\_all} on $\auroc{23}$ throughout.}
\label{tab:app_thomas_mono}
\begin{tabular}{r lll}
\toprule
$k$ & scratch & pre\_head & pre\_all \\
\midrule
5   & 0.708$\pm$0.318 & 0.937$\pm$0.011 & \textbf{1.305$\pm$0.202} \\
10  & 0.795$\pm$0.502 & 0.940$\pm$0.012 & \textbf{1.043$\pm$0.344} \\
15  & 0.999$\pm$0.472 & 0.938$\pm$0.025 & \textbf{1.160$\pm$0.329} \\
20  & 1.167$\pm$0.432 & 0.944$\pm$0.021 & \textbf{1.433$\pm$0.277} \\
25  & 1.232$\pm$0.491 & 0.944$\pm$0.022 & \textbf{1.323$\pm$0.277} \\
30  & 1.371$\pm$0.240 & 0.927$\pm$0.037 & \textbf{1.483$\pm$0.195} \\
35  & 1.349$\pm$0.506 & 0.945$\pm$0.026 & \textbf{1.324$\pm$0.314} \\
40  & \textbf{1.646$\pm$0.192} & 0.934$\pm$0.039 & 1.223$\pm$0.378 \\
45  & 1.340$\pm$0.361 & 0.952$\pm$0.021 & \textbf{1.455$\pm$0.262} \\
50  & 1.513$\pm$0.361 & 0.957$\pm$0.013 & \textbf{1.476$\pm$0.263} \\
55  & 1.409$\pm$0.391 & 0.954$\pm$0.018 & \textbf{1.192$\pm$0.335} \\
60  & 1.267$\pm$0.427 & 0.958$\pm$0.017 & \textbf{1.544$\pm$0.248} \\
65  & 1.483$\pm$0.456 & 0.965$\pm$0.009 & \textbf{1.450$\pm$0.218} \\
70  & 1.501$\pm$0.260 & 0.962$\pm$0.013 & \textbf{1.376$\pm$0.245} \\
75  & \textbf{1.755$\pm$0.139} & 0.968$\pm$0.005 & 1.401$\pm$0.343 \\
80  & 1.477$\pm$0.414 & 0.956$\pm$0.020 & \textbf{1.680$\pm$0.111} \\
85  & 1.487$\pm$0.234 & 0.953$\pm$0.016 & \textbf{1.499$\pm$0.294} \\
90  & \textbf{1.709$\pm$0.153} & 0.963$\pm$0.007 & 1.603$\pm$0.252 \\
95  & \textbf{1.694$\pm$0.225} & 0.951$\pm$0.017 & 1.494$\pm$0.249 \\
100 & \textbf{1.654$\pm$0.152} & 0.949$\pm$0.025 & 1.364$\pm$0.318 \\
\bottomrule
\end{tabular}
\end{table}

\begin{table}[ht]
\centering\small
\caption{Thomas: $\auroc{01}$, $\auroc{12}$, $\auroc{23}$, $\auroc{34}$ (mean\,$\pm$\,std, 5 seeds). \textbf{pre\_all} is the only method that consistently improves $\auroc{23}$ over zero-shot (0.391). \textbf{pre\_head} systematically degrades $\auroc{23}$ below zero-shot across all $k$.}
\label{tab:app_thomas_auroc}
\begin{tabular}{r lll lll lll lll}
\toprule
& \multicolumn{3}{c}{$\auroc{01}$}
& \multicolumn{3}{c}{$\auroc{12}$}
& \multicolumn{3}{c}{$\auroc{23}$}
& \multicolumn{3}{c}{$\auroc{34}$} \\
\cmidrule(lr){2-4}\cmidrule(lr){5-7}\cmidrule(lr){8-10}\cmidrule(lr){11-13}
$k$ & scr & ph & pa & scr & ph & pa & scr & ph & pa & scr & ph & pa \\
\midrule
5   & .607 & .555 & \textbf{.836} & .623 & .540 & \textbf{.711} & .494 & .364 & \textbf{.506} & .763 & .997 & \textbf{.979} \\
10  & .659 & .523 & \textbf{.618} & .595 & .495 & \textbf{.539} & .546 & .380 & \textbf{.557} & .833 & .995 & \textbf{.989} \\
15  & .787 & .536 & \textbf{.736} & .634 & .551 & \textbf{.634} & .524 & .372 & \textbf{.561} & .844 & .996 & \textbf{.977} \\
20  & .753 & .542 & \textbf{.857} & .806 & .541 & \textbf{.697} & .523 & .369 & \textbf{.611} & .819 & .996 & \textbf{.982} \\
25  & .799 & .538 & \textbf{.781} & .712 & .530 & \textbf{.610} & .527 & .390 & \textbf{.607} & .859 & .997 & \textbf{.984} \\
30  & .861 & .537 & \textbf{.920} & .747 & .540 & \textbf{.656} & .530 & .385 & \textbf{.615} & .902 & .997 & \textbf{.959} \\
35  & .839 & .528 & \textbf{.764} & .815 & .514 & \textbf{.579} & .532 & .381 & \textbf{.543} & .884 & .996 & \textbf{.982} \\
40  & .817 & .558 & \textbf{.744} & .825 & .568 & \textbf{.645} & .533 & .383 & \textbf{.650} & .893 & .998 & \textbf{.977} \\
45  & .803 & .555 & \textbf{.820} & .754 & .520 & \textbf{.610} & .539 & .355 & \textbf{.579} & .893 & .997 & \textbf{.987} \\
50  & .875 & .543 & \textbf{.820} & .819 & .539 & \textbf{.573} & .530 & .363 & \textbf{.632} & .904 & .996 & \textbf{.988} \\
55  & .818 & .536 & \textbf{.698} & .707 & .524 & \textbf{.561} & .529 & .374 & \textbf{.571} & .862 & .995 & \textbf{.980} \\
60  & .931 & .557 & \textbf{.863} & .778 & .527 & \textbf{.553} & .507 & .355 & \textbf{.601} & .850 & .997 & \textbf{.970} \\
65  & .812 & .527 & \textbf{.820} & .700 & .521 & \textbf{.603} & .592 & .370 & \textbf{.666} & .897 & .995 & \textbf{.979} \\
70  & .868 & .554 & \textbf{.816} & .707 & .529 & \textbf{.625} & .557 & .363 & \textbf{.605} & .884 & .997 & \textbf{.977} \\
75  & .911 & .545 & \textbf{.731} & .802 & .529 & \textbf{.571} & .519 & .361 & \textbf{.630} & .876 & .996 & \textbf{.985} \\
80  & .882 & .535 & \textbf{.887} & .781 & .533 & \textbf{.611} & .535 & .375 & \textbf{.694} & .884 & .996 & \textbf{.972} \\
85  & .874 & .551 & \textbf{.863} & .741 & .535 & \textbf{.609} & .549 & .360 & \textbf{.601} & .883 & .997 & \textbf{.979} \\
90  & .943 & .546 & \textbf{.881} & .786 & .548 & \textbf{.610} & .525 & .361 & \textbf{.646} & .890 & .996 & \textbf{.977} \\
95  & .966 & .553 & \textbf{.857} & .828 & .556 & \textbf{.625} & .518 & .364 & \textbf{.652} & .907 & .997 & \textbf{.980} \\
100 & .920 & .534 & \textbf{.822} & .786 & .530 & \textbf{.576} & .540 & .376 & \textbf{.602} & .904 & .996 & \textbf{.994} \\
\bottomrule
\end{tabular}
\end{table}

\subsection*{Tigan}
\textbf{Zero-shot:}
$\beta=2.105$,
$\auroc{01}=0.645$,
$\auroc{12}=0.985$,
$\auroc{23}=0.997$,
$\auroc{34}=1.000$.

\begin{table}[ht]
\centering\small
\caption{Tigan: monotonicity slope $\beta$ (mean\,$\pm$\,std, 5 seeds). All pretrained methods match zero-shot from $k=5$; \textbf{pre\_head} alone is sufficient here, confirming that the pretrained representation already covers Tigan's attractor geometry. \textbf{scratch} converges slowly and remains below zero-shot throughout the range.}
\label{tab:app_tigan_mono}
\begin{tabular}{r lll}
\toprule
$k$ & scratch & pre\_head & pre\_all \\
\midrule
5   & 1.157$\pm$0.260 & \textbf{2.058$\pm$0.083} & 2.003$\pm$0.082 \\
10  & 1.445$\pm$0.352 & \textbf{2.035$\pm$0.068} & 2.041$\pm$0.036 \\
15  & 1.502$\pm$0.455 & \textbf{2.086$\pm$0.051} & 2.009$\pm$0.089 \\
20  & 1.682$\pm$0.258 & 1.991$\pm$0.086 & \textbf{2.028$\pm$0.083} \\
25  & 1.885$\pm$0.168 & 1.987$\pm$0.059 & \textbf{1.950$\pm$0.111} \\
30  & 1.838$\pm$0.158 & \textbf{2.018$\pm$0.090} & 2.043$\pm$0.073 \\
35  & 1.724$\pm$0.213 & \textbf{2.000$\pm$0.049} & 1.984$\pm$0.066 \\
40  & 1.965$\pm$0.108 & \textbf{2.008$\pm$0.045} & 1.976$\pm$0.056 \\
45  & 1.818$\pm$0.492 & \textbf{2.024$\pm$0.057} & 2.038$\pm$0.069 \\
50  & 1.817$\pm$0.157 & \textbf{2.015$\pm$0.053} & 2.038$\pm$0.024 \\
55  & 1.757$\pm$0.224 & \textbf{2.069$\pm$0.055} & 2.033$\pm$0.052 \\
60  & 1.723$\pm$0.132 & \textbf{2.021$\pm$0.087} & 2.011$\pm$0.058 \\
65  & 1.720$\pm$0.404 & \textbf{2.066$\pm$0.023} & 2.048$\pm$0.026 \\
70  & 1.782$\pm$0.110 & \textbf{2.011$\pm$0.025} & 1.970$\pm$0.068 \\
75  & 1.942$\pm$0.146 & \textbf{2.031$\pm$0.046} & 2.025$\pm$0.033 \\
80  & 1.865$\pm$0.137 & \textbf{2.009$\pm$0.047} & 2.017$\pm$0.046 \\
85  & 1.917$\pm$0.152 & \textbf{2.053$\pm$0.049} & 2.017$\pm$0.025 \\
90  & 1.838$\pm$0.328 & \textbf{2.015$\pm$0.060} & 2.056$\pm$0.032 \\
95  & 1.859$\pm$0.278 & \textbf{2.058$\pm$0.035} & 2.070$\pm$0.052 \\
100 & 1.829$\pm$0.256 & \textbf{2.045$\pm$0.059} & 2.034$\pm$0.044 \\
\bottomrule
\end{tabular}
\end{table}

\begin{table}[ht]
\centering\small
\caption{Tigan: $\auroc{01}$, $\auroc{12}$, $\auroc{23}$, $\auroc{34}$ (mean\,$\pm$\,std, 5 seeds). $\auroc{12}$, $\auroc{23}$, and $\auroc{34}$ are at ceiling for both pretrained methods from $k=5$. The only boundary where scratch lags meaningfully is $\auroc{01}$.}
\label{tab:app_tigan_auroc}
\begin{tabular}{r lll lll lll lll}
\toprule
& \multicolumn{3}{c}{$\auroc{01}$}
& \multicolumn{3}{c}{$\auroc{12}$}
& \multicolumn{3}{c}{$\auroc{23}$}
& \multicolumn{3}{c}{$\auroc{34}$} \\
\cmidrule(lr){2-4}\cmidrule(lr){5-7}\cmidrule(lr){8-10}\cmidrule(lr){11-13}
$k$ & scr & ph & pa & scr & ph & pa & scr & ph & pa & scr & ph & pa \\
\midrule
5   & .756 & .744 & \textbf{.918} & .902 & .987 & \textbf{.994} & .823 & .998 & \textbf{.995} & .965 & \textbf{1.00} & 1.00 \\
10  & .796 & .824 & \textbf{.931} & .936 & .986 & \textbf{.995} & .785 & .998 & \textbf{.996} & .960 & \textbf{1.00} & 1.00 \\
15  & .848 & .791 & \textbf{.936} & .976 & .986 & \textbf{.994} & .784 & .998 & \textbf{.995} & .997 & \textbf{1.00} & 1.00 \\
20  & .772 & .850 & \textbf{.917} & .885 & .988 & \textbf{.994} & .834 & .998 & \textbf{.995} & .995 & \textbf{1.00} & 1.00 \\
25  & .871 & .882 & \textbf{.930} & .953 & .987 & \textbf{.996} & .788 & .998 & \textbf{.991} & .971 & \textbf{1.00} & .999 \\
30  & .891 & .853 & \textbf{.931} & .976 & .984 & \textbf{.995} & .811 & .998 & \textbf{.994} & .981 & \textbf{1.00} & .997 \\
35  & .814 & .867 & \textbf{.972} & .952 & .983 & \textbf{.990} & .801 & .998 & \textbf{.997} & .994 & \textbf{1.00} & .995 \\
40  & .947 & .872 & \textbf{.941} & .991 & .986 & \textbf{.991} & .779 & .998 & \textbf{.996} & .978 & \textbf{1.00} & .999 \\
45  & .911 & .904 & \textbf{.945} & .977 & .984 & \textbf{.996} & .842 & .998 & \textbf{.994} & .988 & \textbf{1.00} & .993 \\
50  & .943 & .881 & \textbf{.964} & .994 & .985 & \textbf{.992} & .818 & .998 & \textbf{.995} & .987 & \textbf{1.00} & .998 \\
55  & .994 & .886 & \textbf{.949} & .998 & .983 & \textbf{.998} & .774 & .998 & \textbf{.992} & .991 & \textbf{1.00} & .999 \\
60  & .861 & .868 & \textbf{.955} & .944 & .985 & \textbf{.992} & .788 & .998 & \textbf{.996} & .985 & \textbf{1.00} & 1.00 \\
65  & .925 & .854 & \textbf{.951} & .999 & .984 & \textbf{.993} & .834 & .998 & \textbf{.995} & .995 & \textbf{1.00} & 1.00 \\
70  & .872 & .888 & \textbf{.939} & .956 & .986 & \textbf{.992} & .843 & .998 & \textbf{.995} & .986 & \textbf{1.00} & .999 \\
75  & .987 & .878 & \textbf{.959} & .998 & .984 & \textbf{.997} & .800 & .998 & \textbf{.996} & .993 & \textbf{1.00} & .998 \\
80  & .897 & .890 & \textbf{.962} & .972 & .986 & \textbf{.996} & .810 & .998 & \textbf{.995} & .988 & \textbf{1.00} & .998 \\
85  & .915 & .880 & \textbf{.973} & .984 & .984 & \textbf{.997} & .820 & .998 & \textbf{.995} & .995 & \textbf{1.00} & .996 \\
90  & .892 & .892 & \textbf{.961} & .978 & .983 & \textbf{.999} & .795 & .998 & \textbf{.994} & .994 & \textbf{1.00} & .996 \\
95  & .951 & .880 & \textbf{.955} & .982 & .986 & \textbf{.998} & .814 & .998 & \textbf{.995} & .993 & \textbf{1.00} & .998 \\
100 & .908 & .883 & \textbf{.951} & .981 & .985 & \textbf{.998} & .826 & .998 & \textbf{.994} & .993 & \textbf{1.00} & .995 \\
\bottomrule
\end{tabular}
\end{table}

\section{Full Noise Robustness Results}
\label{app:noise}
Table~\ref{tab:noise_full} reports the complete inference-time perturbation 
sweep on the 12 source systems (in-distribution). No retraining is performed; the pretrained GON checkpoint is applied directly.

\paragraph{Perturbation details.}
All perturbations are applied at inference only under \texttt{torch.no\_grad()}. \textbf{Gaussian noise}: SNR is defined on a mean-square power basis, $\mathrm{SNR}_\mathrm{dB} = 10\log_{10}(P_s/P_n)$ where $P_s = \mathrm{mean}(x^2)$
over all channels and timesteps jointly per sample; a single noise power scalar is shared across channels. \textbf{Temporal dropout}: selected timesteps are zero-masked in-place (sequence length preserved), with the same mask applied across all channels simultaneously; each sample draws an independent mask. \textbf{Quantization}: post-training uniform asymmetric quantization applied per channel per sample (min-max over the time axis) in pure PyTorch; zero is a representable level (uniform mid-tread).

\paragraph{Results.}
(1)~Ordinal coherence is preserved down to 20\,dB SNR and falls below $1.0$ between 30 and 20\,dB; at 5\,dB, $\beta=0.216$, but the score remains positively monotone on average.
(2)~Degradation is boundary-asymmetric throughout: $\auroc{34}$ holds at $1.000$ through 30\,dB Gaussian noise and all dropout fractions up to 30\%,
while $\auroc{01}$ falls below $0.5$ at 20\,dB and at 20\% dropout; $\auroc{12}$ is intermediate, tracking $\auroc{01}$ in shape but degrading more slowly.
(3)~Quantization is negligible at all tested precisions: no metric changes by more than $0.002$ between 16-bit and 4-bit, and $\beta$ is unchanged to three decimal places down to 6-bit.

\begin{table}[H]
\centering
\caption{Full noise robustness sweep on the source systems (in-distribution). No retraining; pretrained GON checkpoint applied directly under each condition.}
\label{tab:noise_full}
\small
\begin{tabular}{llccccc}
\toprule
Type & Condition & $\beta$ & $\mathrm{AUROC}_{01}$ & $\mathrm{AUROC}_{12}$ & $\mathrm{AUROC}_{23}$ & $\mathrm{AUROC}_{34}$ \\
\midrule
& clean & 2.065 & 0.965 & 0.993 & 0.994 & 1.000 \\
\midrule
\multirow{5}{*}{Gaussian} & 40\,dB     & 1.946 & 0.932 & 0.944 & 0.997 & 1.000 \\
                          & 30\,dB     & 1.461 & 0.618 & 0.802 & 0.998 & 1.000 \\
                          & 20\,dB     & 0.882 & 0.449 & 0.539 & 0.892 & 0.998 \\
                          & 10\,dB     & 0.522 & 0.428 & 0.480 & 0.680 & 0.824 \\
                          & 5\,dB      & 0.216 & 0.446 & 0.456 & 0.574 & 0.702 \\
\midrule
\multirow{5}{*}{Quantization}
                          & 16-bit     & 2.065 & 0.965 & 0.993 & 0.994 & 1.000 \\
                          & 12-bit     & 2.065 & 0.965 & 0.993 & 0.994 & 1.000 \\
                          & 8-bit      & 2.065 & 0.965 & 0.993 & 0.994 & 1.000 \\
                          & 6-bit      & 2.065 & 0.965 & 0.993 & 0.994 & 1.000 \\
                          & 4-bit      & 2.064 & 0.966 & 0.993 & 0.993 & 1.000 \\
\midrule
\multirow{4}{*}{Dropout}  & 5\%        & 1.702 & 0.727 & 0.881 & 0.993 & 1.000 \\
                          & 10\%       & 1.405 & 0.590 & 0.763 & 0.979 & 1.000 \\
                          & 20\%       & 1.091 & 0.450 & 0.692 & 0.943 & 1.000 \\
                          & 30\%       & 0.877 & 0.372 & 0.636 & 0.896 & 0.999 \\
\bottomrule
\end{tabular}
\end{table}

\end{document}